\documentclass[10pt,journal,compsoc]{IEEEtran}

\usepackage{graphicx}
\usepackage{comment}
\usepackage{epsfig}
\usepackage{amsmath,amssymb} 
\usepackage{color}
\usepackage[table]{xcolor}
\usepackage{tikz}
\usepackage{comment}
\usepackage{multirow}
\usepackage[hang,flushmargin]{footmisc} 
\usepackage{booktabs,tabularx}
\usepackage{enumitem}
\usepackage{makecell}
\usepackage{algorithm}
\usepackage{algorithmic}
\usepackage{threeparttable}
\usepackage{lipsum}
\usepackage[pagebackref=false, breaklinks=true, colorlinks, bookmarks=false]{hyperref}
\usepackage{url}

\usepackage{graphicx}
\usepackage{amsmath}
\usepackage{amssymb}
\usepackage{booktabs}
\usepackage{multirow}
\usepackage{stfloats}
\usepackage{caption}
\usepackage{colortbl}
\usepackage{multirow}
\usepackage{multicol}
 \usepackage{subcaption} 
 \usepackage{pifont}
\usepackage{marvosym}
\definecolor{linecolor}{rgb}{0.82, 0.94, 0.75}

\usepackage{booktabs} 

\definecolor{greenbg}{rgb}{0.9, 1.0, 0.9} 

\newcommand{\tabincell}[2]{\begin{tabular}{@{}#1@{}}#2\end{tabular}}
\newcommand{\revise}[1]{{\color{black} #1}}
\newcommand{\Revise}[1]{{\color{black} #1}}

\definecolor{lowred}{RGB}{238,18,137}

\definecolor{lowerred}{RGB}{255,110,180}

\newcommand{\dplus}[1]{\fontsize{6pt}{0.1em}\selectfont (\textbf{\textcolor{lowred}{#1}})}

\definecolor{rowunit}{RGB}{128,128,255}
\definecolor{evaunit01green}{RGB}{54,125,189}
\newcommand{\evagreen}[1]{\textcolor{evaunit01green}{#1}}
\newcommand{\dtplus}[1]{\fontsize{6pt}{0.1em}\selectfont (\textbf{\evagreen{#1}})}
\makeatletter
\renewcommand{\maketag@@@}[1]{\hbox{\m@th\normalsize\normalfont#1}}%
\makeatother

\ifCLASSOPTIONcompsoc
  \usepackage[nocompress]{cite}
\else
  \usepackage{cite}
\fi

\ifCLASSINFOpdf
\else
\fi
\begin{document}
%
\title{Parameter-Efficient Fine-Tuning in Spectral Domain for Point Cloud Learning}

\author{
Dingkang Liang, Tianrui Feng, Xin Zhou, Yumeng Zhang, Zhikang Zou, Xiang Bai
\IEEEcompsocitemizethanks{

\IEEEcompsocthanksitem Dingkang Liang, Tianrui Feng, Xin Zhou, and Xiang Bai are with Huazhong University of Science and Technology. (dkliang, tianruifeng, xzhou03, xbai)@hust.edu.cn

\IEEEcompsocthanksitem Yumeng Zhang and Zhikang Zou are with Baidu Inc., China.

\IEEEcompsocthanksitem Dingkang Liang, Tianrui Feng, and Xin Zhou make equal contributions. The corresponding author is Xiang Bai (xbai@hust.edu.cn). 

}
}



%
%

\markboth{
IEEE TRANSACTIONS ON PATTERN ANALYSIS AND MACHINE INTELLIGENCE}%
{Shell \MakeLowercase{\textit{et al.}}: Bare Advanced Demo of IEEEtran.cls for IEEE Computer Society Journals}

\IEEEtitleabstractindextext{%
\begin{abstract}

Recently, leveraging pre-training techniques to enhance point cloud models has become a prominent research topic. However, existing approaches typically require full fine-tuning of pre-trained models to achieve satisfactory performance on downstream tasks, which is storage-intensive and computationally demanding. To address this issue, we propose a novel Parameter-Efficient Fine-Tuning (PEFT) method for point cloud, called \textbf{PointGST} (\textbf{Point} cloud \textbf{G}raph \textbf{S}pectral \textbf{T}uning). PointGST freezes the pre-trained model and introduces a lightweight, trainable Point Cloud Spectral Adapter (PCSA) for fine-tuning parameters in the spectral domain. The core idea is built on two observations: 1) The inner tokens from frozen models might present confusion in the spatial domain; 2) Task-specific intrinsic information is important for transferring the general knowledge to the downstream task. Specifically, PointGST transfers the point tokens from the spatial domain to the spectral domain, effectively de-correlating confusion among tokens by using orthogonal components for separation. Moreover, the generated spectral basis involves intrinsic information about the downstream point clouds, enabling more targeted tuning. As a result, PointGST facilitates the efficient transfer of general knowledge to downstream tasks while significantly reducing training costs. 
Extensive experiments on challenging point cloud datasets across various tasks demonstrate that PointGST not only outperforms its fully fine-tuning counterpart but also significantly reduces trainable parameters, making it a promising solution for efficient point cloud learning. 
Moreover, it achieves superior accuracies of 99.48\%, 97.76\%, and 96.18\% on the ScanObjNN OBJ\_BG, OBJ\_ONLY, and PB\_T50\_RS datasets, respectively, establishing a new state-of-the-art, while using only 0.67\% of the trainable parameters. The code will be released at \url{https://github.com/jerryfeng2003/PointGST}.
\end{abstract}

}

\maketitle

\IEEEdisplaynontitleabstractindextext

\IEEEpeerreviewmaketitle

\ifCLASSOPTIONcompsoc
\IEEEraisesectionheading{\section{Introduction}\label{sec:introduction}}
\else
\section{Introduction}
\label{sec:introduction}
\fi

\def\recon{{\scshape ReCon}}

\IEEEPARstart{P}{oint} cloud learning is a fundamental and highly practical task within the computer vision community, drawing significant attention due to its extensive applications such as autonomous driving~\cite{li2023pillarnext,fan2023super}, 3D reconstruction~\cite{melas2023pc2,xiao2023unsupervised}, and embodied intelligence~\cite{xu2023pointllm}, among others. Point clouds are inherently sparse, unordered, and irregular, posing unique challenges to effective analysis.

Nowadays, exploring pre-training techniques to boost the models has become a hot research topic in both natural language processing~\cite{devlin2019bert} and computer vision~\cite{he2022masked}, and has been successfully transferred to the point cloud domain~\cite{pang2022masked,yu2022point,wang2024point}. Representative works like Point-BERT~\cite{yu2022point} and Point-MAE~\cite{pang2022masked} introduce the masked point modeling task for point cloud data, and the subsequent studies~\cite{dong2023act,zhang2022point,chen2024pointgpt,liang2024pointmamba} propose effective customized pre-training tasks. After pre-training, these methods adopt the classical fully fine-tuning (FFT) strategy, where all models' parameters are fine-tuned, leading to notable performance improvements and faster convergence compared to training from scratch. However, the FFT strategy incurs significant GPU memory and storage costs due to updating all parameters of the pre-trained models. With the scaling of models\footnote{Pre-trained point cloud model parameters increased 30$\times$ in three years, from 22.1M in ECCV 2022~\cite{pang2022masked} to 657.2M in ECCV 2024~\cite{qi2024shapellm}.} or increasing needs for fine-tuning enormous new datasets, the demand for storing fine-tuned checkpoints grows, resulting in heightened storage and memory consumption.

The community is well aware of this issue, and to mitigate this challenge, research efforts~\cite{zha2023instance,zhou2024dynamic,tang2024point,fei2024fine} have been directed towards finding a promising fine-tuning strategy, i.e., Parameter-Efficient Fine-Tuning (PEFT). Their common idea is to freeze the pre-trained models and selectively insert extra learnable parameters into the models. The pioneers are IDPT~\cite{zha2023instance} and DAPT~\cite{zhou2024dynamic}, taking an important step toward efficient tuning on point cloud tasks. Specifically, IDPT~\cite{zha2023instance} proposes an instance-aware dynamic prompt tuning to generate a universal prompt for the point cloud data. DAPT~\cite{zhou2024dynamic} generates a dynamic scale for each point token, considering the token significance to the data.

\begin{figure*}[t]
	\begin{center}
		\includegraphics[width=0.96\linewidth]{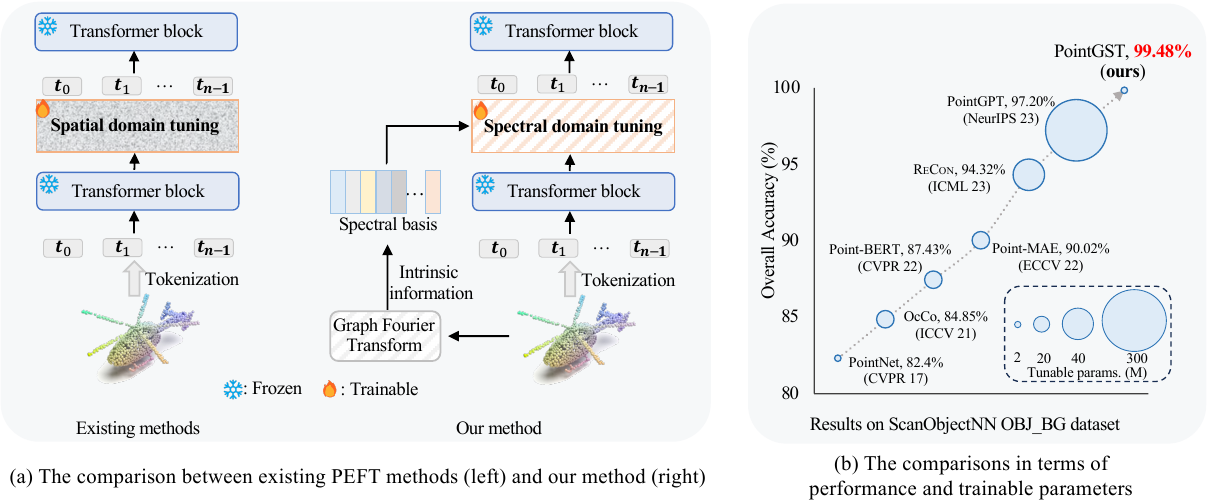}
	\end{center}
    \vspace{-10pt}
	\caption{
(a) Comparison of existing PEFT methods~\cite{zhou2024dynamic,zha2023instance} and our approach. Instead of spatial tuning, we operate in the spectral domain to reduce token confusion and incorporate intrinsic information from downstream point clouds for targeted adaptation. (b) Our PointGST surpasses all prior methods with extremely few trainable parameters, achieving a new record of 99.48\% accuracy on ScanObjectNN OBJ\_BG and exceeding 99\% on this benchmark for the first time.
}
	\label{fig:intro}
\end{figure*}

While the above methods are efficient in training and storage, they still fail to achieve satisfactory performance across various pre-trained models. We attribute this to two factors: \textbf{First}, essentially, these methods are fine-tuned in the spatial domain (Fig.~\ref{fig:intro}(a))\revise{, i.e., the tuning modules process tokens representing localized patches of the point cloud}. In the PEFT setting, pre-training tasks are designed to learn general representations of point clouds, without incorporating prior knowledge from downstream tasks. This leads to the inner tokens (features) from frozen pre-trained models struggling to distinguish the fine-grained structures of the point cloud, where such tokens are referred to as inner confused tokens. \revise{Specifically, even if two parts of the point cloud have similar geometric structures, their output features from the pre-trained model can differ significantly. Current point cloud PEFT methods address this by merging inner confused tokens with newly introduced learnable modules for downstream task adaptation. However, continuously incorporating confusing features from the frozen parameters during the PEFT task can complicate optimization.}

\textbf{Second}, intrinsic structures (e.g., task-specific geometry and relationship of points) of the downstream point clouds are essential for comprehensive analysis~\cite{xiao2023unsupervised,gao2020graphter}. The fixed pre-trained models lack the ability to update their parameters to learn the intrinsic information, relying solely on the representations captured during pre-training. Existing point cloud PEFT methods mainly use the encoded features from the frozen models to generate prompts or feed them into adapters, without explicitly incorporating intrinsic information from downstream point clouds.

To this end, we shift to a fresh perspective: spectral domain fine-tuning. Spectral representations offer significant advantages over spatial domain representations. Specifically, compared to the spatial domain, features in the spectral domain can be easily de-correlated~\cite{rue2005gaussian,zhang2014point}, due to the ability of spectral representations to disentangle complex spatial relationships into distinct frequency components. This de-correlation mitigates the confusion among tokens by using orthogonal components (spectral basis) to separate them. Additionally, the spectral basis obtained by the original point cloud offers direct intrinsic information about the downstream tasks to better guide the fine-tuning process. However, effectively and efficiently fine-tuning the parameters in the spectral domain is not a trivial problem, where two crucial concerns need to be considered: \textbf{1)} what spectral transform to use, and \textbf{2)} how to inject the intrinsic information into the fixed pre-trained model by using this transform. An ideal transform for point clouds should be easy to construct, preserve intrinsic information, and adapt to the irregular structure of point cloud data. We find that the Graph Fourier Transform (GFT) stands out as particularly suitable. \revise{The GFT decomposes confused spatial point tokens into orthogonal graph signals of varying smoothness (the eigenvectors of the Laplace matrix), thus allowing the signal to be separated into orthogonal bases, offering a new perspective for analysis.} Moreover, a properly constructed graph provides a natural representation of the irregular point clouds that are adaptive to their structure, and the graph spectral domain can explicitly reveal geometric structures, from basic shapes to fine details, within a compressed space~\cite{hu2021graph}. 

In this paper, we explore the potential of fine-tuning parameters in the spectral domain. Our approach, illustrated in Fig.~\ref{fig:intro}(a), introduces a novel PEFT method that optimizes a frozen pre-trained \textbf{point} cloud model through \textbf{G}raph \textbf{S}pectral \textbf{T}uning, which we refer to as \textbf{PointGST}. Specifically, we first construct a series of multi-scale point cloud graphs that enable the calculation of global and local spectral basis, which captures intrinsic information such as the geometry of the original point cloud. Then, we propose a lightweight Point Cloud Spectral Adapter (PCSA) to transform the compressed point tokens (serving as graph signal) from the spatial domain to the spectral domain using the spectral basis. Based on the spectral point tokens, our PCSA adopts an effortless shared linear layer to adapt the global and local spectral tokens to meet the downstream tasks. \revise{In the spectral domain, the compressed point tokens are projected onto orthogonal eigenvectors. The goal of spectral domain fine-tuning is to modify the projected coefficients of point tokens within the $n$-dimensional eigenvector space. Due to the bases being orthogonal, the inner confused tokens are de-correlated, thus facilitating the optimization process.} Besides, the information of the generated spectral basis is inherently aware of the unique characteristics of the downstream point clouds, enabling targeted fine-tuning.

Extensive experiments conducted on challenging point cloud datasets across various tasks and data settings demonstrate the effectiveness of our PointGST. For example, our method boosts the large-scale pre-trained models PointGPT-L~\cite{chen2024pointgpt}, only using 0.67\% trainable parameters to achieve 99.48\% accuracy on the ScanObjectNN OBJ\_BG dataset, establishing the new state-of-the-art. To the best of our knowledge, we are the first to push the performance on this dataset to surpass 99\%. We also present the intuitive trend of performance and trainable parameters in Fig.~\ref{fig:intro}(b).

Our main contributions are summarized as follows: \textbf{1)} We propose a novel Parameter-Efficient Fine-Tuning (PEFT) method for point cloud learning, named \textbf{PointGST}, which innovatively fine-tunes parameters from a fresh perspective, i.e., spectral domain. \textbf{2)} PointGST introduces the Point Cloud Spectral Adapter (PCSA) for frozen pre-trained backbones, transferring the inner confused tokens from the spatial domain into the spectral domain. It effectively mitigates the confusion among tokens and brings intrinsic information from downstream point clouds to meet the requirements of the fine-tuned tasks. \textbf{3)} Extensive experiments across various datasets demonstrate that our proposed method outperforms all current point cloud-customized PEFT approaches. Remarkably, we achieve state-of-the-art performance on \revise{eight} challenging point cloud datasets while only using about 0.67\% trainable parameters, serving as a promising option for efficient point cloud fine-tuning.

\section{Related work}

\subsection{Point Cloud Learning}

Constructing structural representations for point clouds is a fundamental problem in computer vision. To address irregularity and sparsity of point clouds, PointNet~\cite{qi2017pointnet} uses shared MLPs for independent feature extraction, while later works integrate local/global information~\cite{qi2017pointnet++,rao2020global,qian2022pointnext,lin2023meta,sun2024x,wang2024gpsformer}, self-attention~\cite{zhao2021point,guo2021pct,wu2022point,wu2024point}, and various regularization strategies~\cite{zhang2023flattening,zhang2023learning,wang2024point}.

Recently, self-supervised pre-training methods~\cite{yu2022point,qi2023contrast,xie2020pointcontrast,liang2024pointmamba,zheng2024point,feng2024shape2scene}, have gained significant attention in the domain for their remarkable transfer capabilities by learning latent representations from unlabeled data and then fine-tuning on various downstream tasks. Point cloud pre-training approaches can be categorized into contrast-based~\cite{xie2020pointcontrast,afham2022crosspoint,chen2023clip2scene} and reconstruct-based~\cite{yu2022point,pang2022masked,zhang2022point,dong2023act,chen2024pointgpt,qi2024shapellm} paradigms. In the contrast-based paradigm, PointContrast~\cite{xie2020pointcontrast} and CrossPoint~\cite{afham2022crosspoint} extract latent information from different views of the point cloud. Conversely, reconstruction-based methods involve randomly masking point clouds and then using autoencoders to reconstruct the original input. Among them, Point-BERT~\cite{yu2022point} learns by comparing masked encodings with outputs from a dVAE-based point cloud tokenizer. Point-MAE~\cite{pang2022masked} and Point-M2AE~\cite{zhang2022point} reconstruct original point clouds via an autoencoder. Following this, PointGPT~\cite{chen2024pointgpt} leverages an extractor-generator-based transformer as a GPT-style approach.
To tackle data scarcity in 3D representation learning, interest in using cross-modal data for point cloud analysis is growing. For instance, ACT~\cite{dong2023act} trains 3D representation models using cross-modal pre-trained models as teachers, and \recon~\cite{qi2023contrast,qi2024shapellm} unifies reconstruction and cross-modal contrast modeling further.

Most pre-trained models are transferred to downstream 3D tasks via full fine-tuning. However, with the growing scale of models~\cite{chen2024pointgpt,qi2024shapellm}, full fine-tuning incurs substantial tuning and storage costs and may dilute pre-trained knowledge. This work aims to address these issues through efficient transfer learning algorithms.

\subsection{Parameter-Efficient Fine-Tuning}

Parameter-Efficient Fine-Tuning (PEFT) introduces small trainable modules to efficiently adapt large models and has attracted significant attention in both NLP~\cite{chen2022adaptformer,sung2022lst,zhang2023adaptive,shi2024dept,ding2023parameter} and computer vision~\cite{jia2022visual,lian2022scaling,zaken2022bitfit,tu2023visual}. Mainstream PEFT methods are typically Adapter-based~\cite{houlsby2019parameter,hu2021lora,chen2022adaptformer,sung2022lst,li2024adapter}, which insert adapters between frozen layers or Prompt-based~\cite{lester2021power,li2021prefix,jia2022visual,shi2024dept}, which prepend learnable tokens to model inputs. Other approaches include feature modulation~\cite{lian2022scaling} and bias tuning~\cite{zaken2022bitfit}.

Recently, some works~\cite{zha2023instance,tang2024point,zhou2024dynamic,fei2024fine,li2024adapt,tang2024any2point,sun24ppt,fei2024parameter} attempt to explore PEFT in point cloud tasks.
As the earliest method, IDPT~\cite{zha2023instance} applies DGCNN~\cite{wang2019dynamic} to extract dynamic prompts for different instances instead of the traditional static prompts. Point-PEFT~\cite{tang2024point} attempts to aggregate local point information during fine-tuning. DA~\cite{fei2024fine} introduces a dynamic aggregation strategy to replace previous static aggregation like mean or max pooling for pre-trained point cloud models. DAPT~\cite{zhou2024dynamic} combines the dynamic scale adapters with internal prompts as an efficient way of point cloud transfer learning.

These point cloud methods operate in the spatial domain, where tuning modules handle tokens representing localized patches of the point cloud, each corresponding to a specific point region. We argue that the pre-training tasks prioritize general point cloud representations without prior downstream tasks, leading to frozen pre-trained models struggling to distinguish fine-grained point cloud structures in the spatial domain.

\subsection{Spectral Methods for Point Clouds}

Spectral-based methods have achieved significant progress in point cloud understanding~\cite{te2018rgcnn,chen2018pointagcn,li2019deepgcns,lu2020pointngcnn}, either by modeling local details for fine-grained feature extraction~\cite{rao2020global,wen2024pointwavelet} or applying graph spectral convolutions for compact representations~\cite{te2018rgcnn,chen2018pointagcn,li2019deepgcns,lu2020pointngcnn}. For example, Wang et al.\cite{wang2018local} combine spectral feature learning and clustering to address limitations of local aggregation, while Zhang et al.\cite{zhang2020hypergraph} estimate hypergraph spectral components in various settings. Other works include spectral domain attacks~\cite{liu2023point}, efficient wavelet-based spectral transforms~\cite{wen2024pointwavelet}, and spectral-domain GANs for 3D shape generation~\cite{ramasinghe2020spectral}.

Distinct from prior spectral approaches, this paper focuses on efficient parameter-efficient fine-tuning (PEFT). We propose PointGST, which fine-tunes pre-trained point cloud models in the spectral domain, differing in both motivation and technique from previous spectral-based methods.

\section{Preliminary}
This section revisits the parameter-efficient fine-tuning paradigm and the concept of Graph Fourier Transform.
\subsection{Parameter-Efficient Fine-Tuning Paradigm}

Given a downstream training set $\boldsymbol{\Gamma}\left(\boldsymbol{x};\boldsymbol{y}\right )$ ($\boldsymbol{x}$ is the input and $\boldsymbol{y}$ is the label), parameter-efficient fine-tuning (PEFT) provides a practical solution by adapting pre-trained models $\mathcal{F}$ to downstream tasks in a more efficient manner. Existing PEFT methods focus on tuning only a small subset of parameters $\boldsymbol{\theta}'$. The optimized parameter $\boldsymbol{\theta}'^{*}$ can be represented as:
\begin{equation}
   \boldsymbol{\theta}'^{*}=\mathop{\arg\min}\limits_{\boldsymbol{\theta}'} \ell \left ( \mathcal{F} \left(\boldsymbol{x};\boldsymbol{\theta}, \boldsymbol{\theta}'\right ),\boldsymbol{y} \right ), \left | \boldsymbol{\theta}' \right | \ll \left | \boldsymbol{\theta} \right |,
\end{equation}
where $\ell$ represents the loss function for downstream tasks. The number of parameters in $\boldsymbol{\theta}'^{*}$, denoted $\left | \boldsymbol{\theta}'^{*} \right |$, equals $\left | \boldsymbol{\theta}' \right |$ and is significantly smaller than the pre-trained $\boldsymbol{\theta}$. During fine-tuning, the pre-trained $\boldsymbol{\theta}$ remains fixed while only $\boldsymbol{\theta}'$ is updated. Techniques for incorporating $\boldsymbol{\theta}'$ include adding lightweight additional parameters and selecting a small subset of $\boldsymbol{\theta}$. 

Adapter-like methods~\cite{chen2022adaptformer,houlsby2019parameter,yu2024visual}, adding lightweight extra parameters, are typical of the PEFT paradigm. Specifically, an Adapter-based approach usually includes a down-sampling layer $\boldsymbol{W}_{d}$ to reduce the feature dimension $C$ to $r$, an activation function $\Phi$, and an up-sampling layer $\boldsymbol{W}_{u}$ to recover the dimension. The low-rank dimension $r~(r\ll C)$ serves as a hyper-parameter for adapters. The formulation can be expressed as:
\begin{equation}
\hat{\boldsymbol{x}}=\Phi( \boldsymbol{x}\boldsymbol{W}_{d}^{\top})\boldsymbol{W}_{u}^{\top},
\end{equation}
where $\boldsymbol{x}\in\mathbb{R}^{n\times C}$ represents the input vectors from a pre-trained model. The output $\hat{\boldsymbol{x}}$ is added back to the original computation graph, with multiple Adapter structures comprising the new parameters $\boldsymbol{\theta}'$.

\subsection{Graph Fourier Transform}

Let us define a graph $\mathcal{G} = \{ \mathcal{V}, \mathcal{E}, \boldsymbol{\mathcal{W}} \}$, where $\mathcal{V}$ is the set of vertices ($|\mathcal{V}| = n$), $\mathcal{E}$ represents the edges, and $\boldsymbol{\mathcal{W}}$ is the adjacency matrix. In our context, a graph signal $\boldsymbol{z} \in \mathbb{R}^{n\times c}$ assigns a vector ($\boldsymbol{z}_{i} \in \mathbb{R}^{1\times c}$) to each vertex, while the entry $w_{i,j}$ in $\boldsymbol{\mathcal{W}}$ indicates the weight of the edge between vertices $i$ and $j$, reflecting their similarity. Then the Laplacian matrix~\cite{shuman2013emerging} is defined as $\boldsymbol{L} = \boldsymbol{D} - \boldsymbol{\mathcal{W}}$, where $\boldsymbol{D}$ is a diagonal matrix with each element $d_{i,i} = \sum_{j=0}^{n-1} w_{i,j}$ representing the degree of vertex $i$. For an undirected graph with non-negative weights, $\boldsymbol{L}\in \mathbb{R}^{n \times n}$ is real, symmetric, and positive semi-definite~\cite{chung1997spectral}. This allows for the eigen-decomposition $\boldsymbol{L} = \boldsymbol{U} \boldsymbol{\Lambda} \boldsymbol{U}^\top$, where $\boldsymbol{U} = [\mathbf{u}_0, \ldots, \mathbf{u}_{n-1}]$ is an orthonormal matrix of eigenvectors $\mathbf{u}_i$, and the eigenvalues $\boldsymbol{\Lambda} = \mathrm{diag}(\lambda_0, \ldots, \lambda_{n-1})$ are often interpreted as the graph frequencies.

Specifically, the eigenvectors above can be viewed as a set of orthogonal spectral basis, allowing the Graph Fourier Transform (GFT) of the graph signal $\boldsymbol{z}$ to be expressed as:
\begin{equation}
    \hat{\boldsymbol{z}} = \mathtt{GFT}(\boldsymbol{z}) = \boldsymbol{U}^\top \boldsymbol{z},
\end{equation}
where $\hat{\boldsymbol{z}} \in \mathbb{R}^{n\times c}$ represents the spectral coefficients for different graph frequencies. Similarly, the inverse Graph Fourier Transform (iGFT) is then defined as:
\begin{equation}
\boldsymbol{z} = \mathtt{iGFT}(\hat{\boldsymbol{z}}) = \boldsymbol{U} \hat{\boldsymbol{z}}.
\end{equation}

Our method associates graph vertices with key points in the point cloud, constructing the adjacency matrix $\boldsymbol{\mathcal{W}}$ considering the distances among these points.

\begin{figure*}[t]
	\begin{center}
		\includegraphics[width=1.0\linewidth]{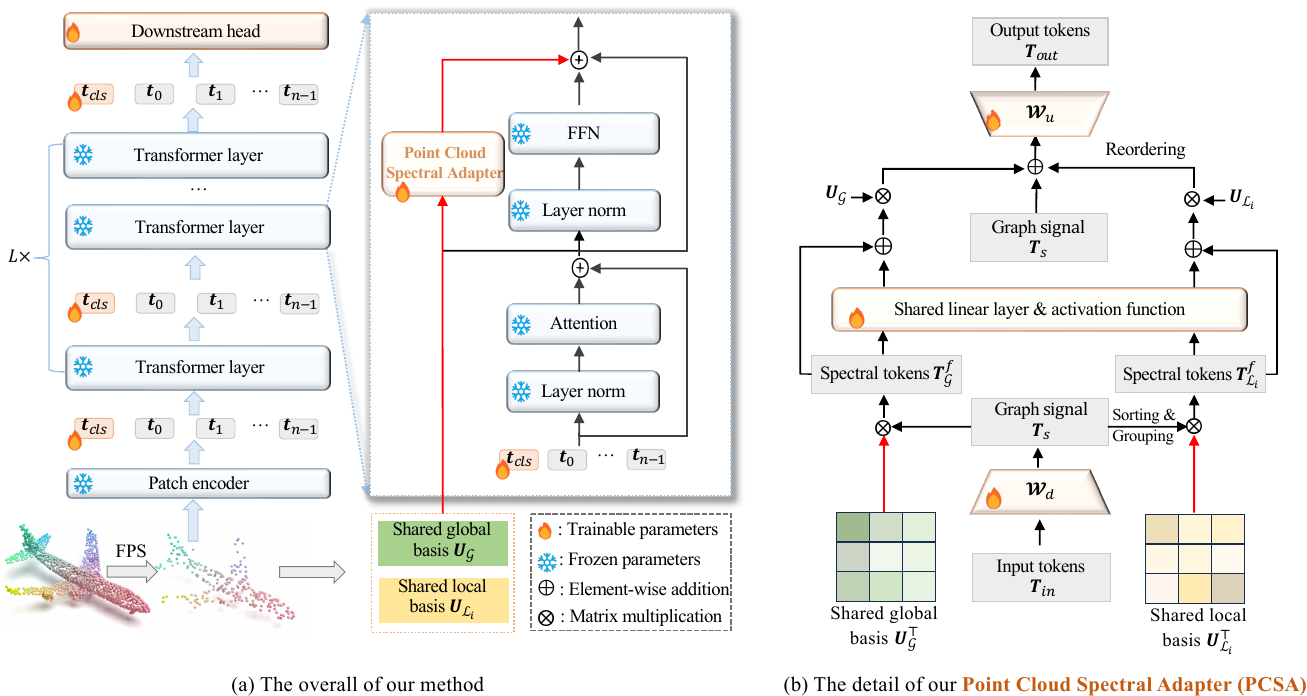}
	\end{center}
        \vspace{-10pt}
	\caption{
(a) The overall of our PointGST. During the fine-tuning phase, we freeze the given pre-trained backbones and only fine-tune the proposed lightweight Point Cloud Spectral Adapter (PCSA). (b) The details of PCSA. We treat the point tokens as graph signals and then transfer the point tokens from the spatial domain to the spectral domain for tuning. 
}
	\label{fig:pipeline}
\end{figure*}

\section{Method}

Fully fine-tuning pre-trained point cloud models yields strong performance but incurs substantial training costs. To address this challenge, we propose a novel method called PointGST, where the overall framework is illustrated in Fig.~\ref{fig:pipeline}. Specifically, our method mainly consists of an off-the-shelf pre-trained backbone and a series of lightweight Point Cloud Spectral Adapters (PCSA) integrated into each Transformer layer. During the fine-tuning, the entire backbone is frozen, and only a small set of trainable parameters (i.e., PCSA) are used to capture task-specific knowledge. The PCSA transforms the point tokens from the spatial to the spectral domain, which de-correlates the inherent confusion among the point tokens from the frozen backbone. It efficiently fine-tunes fewer parameters for downstream tasks while introducing intrinsic information from the downstream point clouds for targeted tuning. As a result, it can significantly reduce the training cost and effectively connect the general knowledge from the pre-training tasks with downstream tasks, leading to promising performance.

\subsection{Transformer Encoder}
\label{sec:transformer_encoder}

Due to its flexible scalability, the Transformer encoder has become the predominant backbone for pre-training in point cloud analysis~\cite{chen2024pointgpt,pang2022masked,qi2023contrast,dong2023act,yu2022point}. Specifically, after pre-processing steps such as Farthest Point Sampling (FPS) and Grouping, a lightweight PointNet is used to generate a series of point tokens ($\left \{ \boldsymbol{t}_i \in  \mathbb{R}^{1\times d} \mid 0  \le i \le n-1\right \} 
$), where $d$ is the embedding dimension and $n$ indicates the number of point patches. A classification token ($\boldsymbol{t}_{cls} \in \mathbb{R}^{1\times d}$) is then concatenated with these point tokens, forming $\boldsymbol{T}_0 \in \mathbb{R}^{(n+1)\times d}$, which serves as the input to an $L$-layer Transformer. Each Transformer layer comprises an $\mathtt{Attention}$ module and a Feed-Forward Network ($\mathtt{FFN}$), which are responsible for extracting token-to-token and channel-wise information, respectively:

\begin{equation} \begin{aligned} \boldsymbol{T}^ {\prime}_i & = \mathtt{Attention}\left (\mathtt{LN}\left (\boldsymbol{T}_{i-1}\right) \right) + \boldsymbol{T}_{i-1},\\ 
\boldsymbol{T}_i & = \mathtt{FFN}\left (\mathtt{LN}\left (\boldsymbol{T}^ {\prime}_i\right )\right ) + \boldsymbol{T}^ {\prime}_i,
\end{aligned} \end{equation}
where $\boldsymbol{T}_i\in \mathbb{R}^{(n+1)\times d}$ represents the output of the $i$-th layer, and $\mathtt{LN}$ denotes layer normalization. The $\mathtt{Attention}$ module and $\mathtt{FFN}$ are the most computationally and parameter-intensive components of the Transformer. In our approach, we keep the entire Transformer encoder frozen and inject the proposed learnable Point Cloud Spectral Adapter (PCSA) into the $\mathtt{FFN}$ in a parallel configuration.

\subsection{Efficient Fine-Tuning in Spectral Domain}
\label{sec:key_design}

Compared with the spatial domain, the point tokens in the spectral domain can be easily de-correlated by distinct frequency components. Fine-tuning the parameters in the spectral domain mitigates the confusion among tokens by using an orthogonal basis to separate them and provides compact intrinsic information from downstream point clouds. To achieve this, we introduce a simple and lightweight Point Cloud Spectral Adapter (PCSA) designed to learn task-specific knowledge in the spectral domain, as illustrated in Fig.~\ref{fig:pipeline}(b). The spectral fine-tuning process is as follows:

\begin{itemize}
    \item First, we transform the sampled $n$ key points (for each point patch and point token) into multi-scale point cloud graphs, consisting of one global graph and $k$ local sub-graphs. Subsequently, we generate the global Laplacian matrix $\boldsymbol{L}_\mathcal{G}$ and the local Laplacian matrices $\boldsymbol{L}_\mathcal{L}$.
    By performing eigenvalue decomposition on $\boldsymbol{L}_\mathcal{G}$ and $\boldsymbol{L}_\mathcal{L}$, we obtain two types of eigenvector matrix, termed $\boldsymbol{U}_\mathcal{G}$ and $\boldsymbol{U}_\mathcal{L}$, respectively.

    \item Next, these eigenvector matrices ($\boldsymbol{U}_\mathcal{G}$ and $\boldsymbol{U}_\mathcal{L}$) are shared across all frozen Transformer layers as the spectral basis for the Graph Fourier Transform (GFT). At each transformer layer, GFT is performed on the low-rank point tokens (serving as the graph signal) $\boldsymbol{T}_{s}$ with $\boldsymbol{U}_\mathcal{G}$ and $\boldsymbol{U}_\mathcal{L}$ to obtain global and local spectral domain tokens, respectively.

    \item Then, the PCSA uses a shared linear layer to adapt the features. Following this, we recover the tokens to the spatial domain via inverse Graph Fourier Transform (iGFT) to align with the output of the transformer layer. 
    
\end{itemize}

In the following contents, we elaborately describe how to fine-tune the parameters in the spectral domain.

\subsubsection{From Point Cloud to Graph}
\label{sec:point2graph}
To implement spectral tuning, we first opt to convert the point cloud into a customized point graph $\mathcal{G}= \{ \mathcal{V}, \mathcal{E}, \boldsymbol{\mathcal{W}} \}$ with vertices $\mathcal{V}$, edges $\mathcal{E}$ and adjacency matrix $\boldsymbol{\mathcal{W}} \in \mathbb{R}^{n\times n}$. An ideal $\mathcal{G}$ should: \textbf{1)} Contain $n$ vertices ($\left | \mathcal{V} \right | =n $) representing the $n$ essential points in the point cloud. \textbf{2)} The element $w_{i,j}$ in $\boldsymbol{\mathcal{W}}$ quantifies the relationship between $i$-th and $j$-th points, where the highest self-correlation is reflected by the diagonal element. \textbf{3)} The relationship between points is influenced by their geometric properties, with closer point clouds exhibiting a stronger relationship.

Considering the design principles above, we calculate the relationship between the key points in a point patch. Then, we measure their pairwise distance to reflect the intrinsic information and capture the underlying structure. Specifically, we compute a point-pairwise distance matrix $\boldsymbol{\Delta}\in \mathbb{R}^{n\times n}$, where each element $\delta_{i,j}\in \mathbb{R}$ represents the Euclidean distance between $i$-th and $j$-th points. However, such a distance matrix fails to meet principles \textbf{2)} and \textbf{3)}, e.g., the distance from the $i$-th point to itself is zero. Therefore, we further propose a simple data-dependent scaling strategy to define the adjacency matrix $\boldsymbol{\mathcal{W}}$, and the element for $i$-th row and $j$-th column as follows:
\begin{equation}
\begin{aligned}
w_{i,j} = \frac{1}{\frac{\delta_{i,j}}{\min(\boldsymbol{\Delta})} + \boldsymbol{I}_{i,j}},
\end{aligned}
\end{equation}
where $\min(\boldsymbol{\Delta}) =  \min(\delta_{i,j}, i \neq j)$ is the minimum value in $\boldsymbol{\Delta}$ excluding zero, which varies across different samples and is data-dependent. $\boldsymbol{I}\in \mathbb{R}^{n\times n}$ is an identity matrix. Using the geometric locations of key points and a data-dependent scaling strategy, we construct a graph with weights based on a distance metric between the points, incorporating all the desired properties.

\begin{figure}[t]
	\begin{center}
		\includegraphics[width=0.96\linewidth]{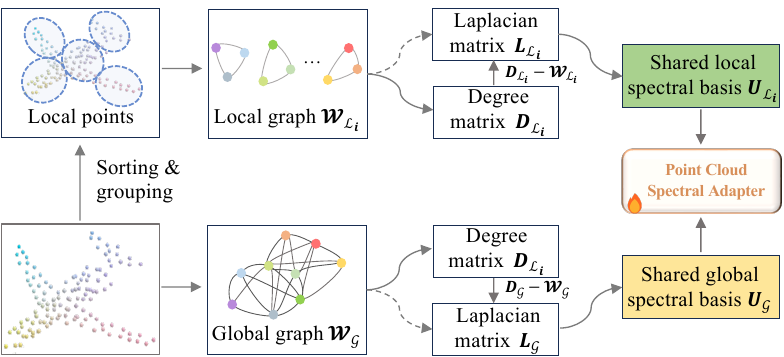}
	\end{center}
    \vspace{-5pt}
	\caption{The detailed process of global and local spectral basis generation. }
	\label{fig:basis_generation}
\end{figure}

\subsubsection{Multi-Scale Point Cloud Graphs}
\label{sec:hierarchical-group}

To comprehensively capture the underlying structure in point clouds, we generate a series of graphs that enable the calculation of both global and local eigenvectors, as shown in Fig.~\ref{fig:basis_generation}. Such an approach ensures a thorough usage of spectral features across different scales. 

\textbf{Global graph.} We first treat all key points as an entirety and transform them into a global graph\footnote{For simplicity, we utilize the adjacency matrix to represent a graph.} $\boldsymbol{\mathcal{W}}_\mathcal{G}$. The details for translating the original point cloud to a graph are detailed in Sec.~\ref{sec:point2graph}. 

\textbf{Local graph.} We then propose constructing local sub-graphs from neighboring point clouds to capture the local structure by rearranging the $n$ key points into $k$ groups along a particular pattern. To be specific, the original key point set $\left \{ p_{0},\cdots,p_{n-1}\right \}$ is scanned and sorted into $\left \{ p'_{0},\cdots,p'_{n-1}\right \}$. This sorted set is then evenly divided into $k$ groups, with each group containing $m$ points, making $n = k\times m$. The $i$-th local point group can be defined as $\mathcal{L}_i = \left \{ p'_{im},\cdots,p'_{(i+1)m-1} \mid 0\le i\le k-1\right \}$. Based on this, we construct local sub-graphs for each group, resulting in $\{ \boldsymbol{\mathcal{W}}_{\mathcal{L}_0},\cdots,\boldsymbol{\mathcal{W}}_{\mathcal{L}_i},\cdots\boldsymbol{\mathcal{W}}_{\mathcal{L}_{k-1}} \mid 0\le i\le k-1 \}$.

Both constructed global and local graphs effectively retain the intrinsic information of the original point cloud and underlying structure. These graphs subsequently will be used to generate the spectral basis, as detailed in the following subsection. Besides, the construction processing only needs to be implemented once for efficiency, i.e., the generated local and global graphs will be shared for all transformer layers.

\subsubsection{Point Cloud Spectral Adapter}

The proposed Point Cloud Spectral Adapter (PCSA) aims to convert the point tokens from the spatial to the spectral domain, and then effectively fine-tune the transformed signal, as shown in Fig.~\ref{fig:pipeline}(b). Our PCSA is simple and extremely lightweight, as it just contains two matrices used for dimension adjustment and one shared linear layer used to fine-tune the spectral point tokens.

\textbf{Down-projection.} Let us denote the input tokens as $\boldsymbol{T}_{in}\in \mathbb{R}^{n\times C}$, where $C$ is the input dimension. Here, we omit the class token and refer to $\boldsymbol{T}_{in}$ as the $n$ point tokens in this section. Our PCSA begins with a downward projection using trainable parameters $\boldsymbol{W}_d \in \mathbb{R}^{r \times C}$, with $r\ll C$. The low-rank graph signal $\boldsymbol{T}_{s}$ is obtained by:
\begin{equation}
\boldsymbol{T}_{s}=\boldsymbol{T}_{in}\boldsymbol{W}_{d}^{\top},
\end{equation}
where $\boldsymbol{T}_{s}\in \mathbb{R}^{n\times r}$ provides a general representation of point patches. The $n$ point patches are embedded by the corresponding $n$ key points with their neighbors, enabling $\boldsymbol{T}_{s}$ to function as a graph signal with the key points as vertices. The $\boldsymbol{T}_{s}$ is also rearranged into $k$ local graph signal, as mentioned in Sec.~\ref{sec:hierarchical-group}, resulting in $\boldsymbol{T}_{s}^{\mathcal{L}_{i}}\in \mathbb{R}^{m \times r} (0\le i \le k-1)$, corresponding to the key points in $\mathcal{L}_{i}$. Also, the $\boldsymbol{T}_{s}$ can serve as the global graph signal notated as $\boldsymbol{T}_{s}^{\mathcal{G}}\in \mathbb{R}^{n \times r}$.

\textbf{Spectral basis generation.} After the down-projection, the graph signal $\boldsymbol{T}_{s}$ still remains in the spatial domain. 
Two preparations are required before applying Graph Fourier Transform (GFT) in the spatial point clouds, first to compute the Laplacian matrix $\boldsymbol{L}_\mathcal{G}\in \mathbb{R}^{n\times n}$ and $\boldsymbol{L}_{\mathcal{L}_i}\in \mathbb{R}^{m \times m}$ for the constructed global graph $\boldsymbol{\mathcal{W}}_{\mathcal{G}}\in \mathbb{R}^{n\times n}$ and local sub-graphs $\boldsymbol{\mathcal{W}}_{\mathcal{L}_i}\in \mathbb{R}^{m \times m}$, respectively. We then perform eigenvalue decomposition to generate the spectral basis for the GFT and its inverse, as shown in Fig.~\ref{fig:basis_generation}. 

For simplicity, the following content uses the global spectral for illustration. 
Specifically, we first produce the diagonal degree matrix $\boldsymbol{D}_\mathcal{G}$ for the global graph $\boldsymbol{\mathcal{W}}_{\mathcal{G}}$:
\begin{equation}
\begin{aligned}
d_{i,j}=\begin{array}{l} 
  \left\{\begin{matrix} 
   {\sum_{l=0}^{n-1}} w_{i,l}&,i=j\\ 
  0&,i\neq j
\end{matrix}\right.
\end{array},
\end{aligned}
\end{equation}
where each entry $d_{i,i}$ indicates the sum of the weights on connected edges for each node (key point). Based on $\boldsymbol{D}_\mathcal{G}$, the graph Laplacian matrix is represented as $\boldsymbol{L}_\mathcal{G}=\boldsymbol{D}_\mathcal{G} - \boldsymbol{\mathcal{W}}_\mathcal{G}$.
The constructed graphs contain real and non-negative edge weights, making $\boldsymbol{L}_\mathcal{G}$ a real, symmetric, and positive semi-deﬁnite matrix~\cite{chung1997spectral}. Therefore, $\boldsymbol{L}_\mathcal{G}$ can be decomposed as:
\begin{equation}
\boldsymbol{L}_\mathcal{G} = \boldsymbol{U}_\mathcal{G} \boldsymbol{\Lambda}_\mathcal{G} \boldsymbol{U}^{\top}_\mathcal{G},
\end{equation}
 where $\boldsymbol{U}_\mathcal{G} = \left [ \mathbf{u}_0, \cdots, \mathbf{u}_{n-1} \right ] \in \mathbb{R}^{n\times n}$ is an orthogonal matrix of eigenvectors $\mathbf{u}_i$, formed as a set of spectral basis for spectral domain. Similarly, we can also obtain the Laplacian matrix $\boldsymbol{L}_{\mathcal{L}_i}$ and $\boldsymbol{U}_{\mathcal{L}_i}\in \mathbb{R}^{m \times m}$ for the local sub-graph $\boldsymbol{\mathcal{W}}_{\mathcal{L}_i}$. The $\boldsymbol{U}_\mathcal{G}$ and $\boldsymbol{U}_{\mathcal{L}_i}$ are computed only once and shared across all transformer layers.

\textbf{Graph Fourier Transform.} We then perform a matrix multiplication of low-dimensional token (graph signal) $\boldsymbol{T}_{s}$ with spectral basis $\boldsymbol{U}_\mathcal{G}$ and $\boldsymbol{U}_{\mathcal{L}_i}$ to represent the graph signal in the spectral domain. Specifically, for the global 
graph signal $\boldsymbol{T}_{s}$, we apply GFT by multiplying it with the global basis $\boldsymbol{U}_\mathcal{G}$ to output $\boldsymbol{T}_\mathcal{G}^{f}$, derive global spectral information. Similarly, the $i$-th local graph signal $\boldsymbol{T}_{s}^{\mathcal{L}_i}$ will multiply with the local basis $\boldsymbol{U}_{\mathcal{L}_i}$, resulting in $\boldsymbol{T}_{\mathcal{L}_i}^{f} $:
\begin{equation}
\begin{array}{c}
\label{eq:gft-method}
\hspace{-3.75mm}
\resizebox{.88\hsize}{!}{$\boldsymbol{T}_\mathcal{G}^{f}=\mathtt{GFT}(\boldsymbol{T}_{s}) = \boldsymbol{U}_\mathcal{G}^{\top}\boldsymbol{T}_{s},~~\boldsymbol{T}_{\mathcal{L}_i}^{f}=\mathtt{GFT}(\boldsymbol{T}_{s}^{\mathcal{L}_i}) = \boldsymbol{U}_{\mathcal{L}_i}^{\top}\boldsymbol{T}_{s}^{\mathcal{L}_i}.$}
\end{array}
\end{equation}
Next, all tokens are passed through a shared linear layer featuring a residual connection:
\begin{equation}
\begin{array}{c}
\hspace{-3.75mm}
\resizebox{.9\hsize}{!}{$\boldsymbol{T}_\mathcal{G}^{f'} = \boldsymbol{T}_\mathcal{G}^{f} + \mathtt{act}(\mathtt{Linear}(\boldsymbol{T}_\mathcal{G}^{f})),\,\boldsymbol{T}_{\mathcal{L}_i}^{f'} = \boldsymbol{T}_{\mathcal{L}_i}^{f} + \mathtt{act}(\mathtt{Linear}(\boldsymbol{T}_{\mathcal{L}_i}^{f})),$}
\end{array}
\end{equation}
where the $\mathtt{act}$ refers to the Swish activation function~\cite{ramachandran2017searching}. We initialize the linear layer to zero for stable training.

\textbf{Inverse Graph Fourier Transform.} To align with the output of the transformer layer, these tuned tokens $\boldsymbol{T}_{\mathcal{L}_i}^{f'} $ and $\boldsymbol{T}_\mathcal{G}^{f'} $ in the spectral domain will be further transformed back to the spatial domain via the inverse GFT (iGFT):
\begin{equation}
\begin{array}{c}
\label{eq:igft-method}
\hspace{-3.75mm}
\resizebox{.9\hsize}{!}{$\hat{\boldsymbol{T}}_\mathcal{G}^{f} = \mathtt{iGFT}(\boldsymbol{T}_\mathcal{G}^{f'}) = \boldsymbol{U}_\mathcal{G}\boldsymbol{T}_\mathcal{G}^{f'},~~\hat{\boldsymbol{T}}^{f}_{\mathcal{L}_i} = \mathtt{iGFT}(\boldsymbol{T}^{f'}_{\mathcal{L}_i}) = \boldsymbol{U}_{\mathcal{L}_i}\boldsymbol{T}^{f'}_{\mathcal{L}_i},$}
\end{array}
\end{equation}
where both $\hat{\boldsymbol{T}}_\mathcal{G}^{f}$ and $\hat{\boldsymbol{T}}^{f}_{\mathcal{L}_i}$ are the converted spatial tokens.
It should be noted that due to the introduced sorting and grouping strategy, the tokens of local spectral represent point cloud patches that do not align with the original token order in the backbone's input. Therefore, it is necessary to reorder the spatial local tokens to match the order of $\boldsymbol{T}_{s}$. Specifically, we concatenate the $\boldsymbol{T}_{\mathcal{L}_i}^{f}$ of all sub-graphs and reordered it, resulting in the tokens $\hat{\boldsymbol{T}}_{\mathcal{L}}^{f} \in \mathbb{R}^{n\times r}$:
\begin{equation}
\begin{aligned}
\hat{\boldsymbol{T}}_{\mathcal{L}}^{f}  = \text{reorder}(\text{concat}[\hat{\boldsymbol{T}}_{\mathcal{L}_0}^{f}, \dots, \hat{\boldsymbol{T}}_{\mathcal{L}_{k-1}}^{f} ]).
\end{aligned}
\end{equation}

\textbf{Up-projection and output.} The PCSA finally employs a zero-initialized up-projection matrix $\boldsymbol{W}_{u} \in \mathbb{R}^{C \times r}$ and a scale $s\in \mathbb{R}$ to restore the fine-tuned compressed representations to the dimensions processed by the backbone: 
\begin{equation}
\begin{aligned}
\boldsymbol{T}_{out}&=s\times(\mathtt{act}(\boldsymbol{T}_{s}) + \hat{\boldsymbol{T}}_\mathcal{G}^{f} + \hat{\boldsymbol{T}}_{\mathcal{L}}^{f})\boldsymbol{W}_{u}^{\top},
\end{aligned}
\end{equation}
where the output of the proposed point cloud spectral adapter $\boldsymbol{T}_{out}$ is added back to the $\mathtt{FFN}$ of the transformer block, as shown in Fig.~\ref{fig:pipeline}(a).

Through the PCSA, we successfully fine-tune the parameter in the spectral domain, effectively transferring the general knowledge to the downstream tasks.

\begin{figure}[t]
        \centering
	\includegraphics[width=0.99\linewidth]{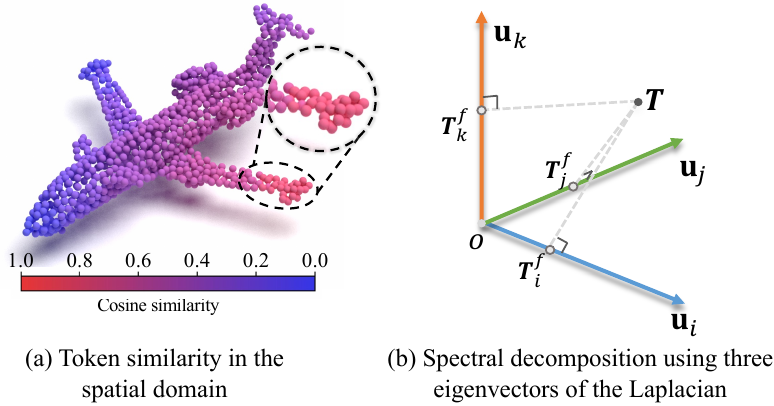}
        \vspace{-5pt}
	\caption{
    \revise{An intuitive overview of spatial and spectral domains. (a) Tokens in the spatial domain after Point-MAE~\cite{pang2022masked} pre-training, where red/blue indicate high/low cosine similarities to the zoomed-in area, respectively. (b) The spectral decomposition projects point tokens $\boldsymbol{T}$ on $n$ eigenvector of the Laplacian (using three as an example).}
    }
	\label{fig:spatial_and_spectral_tokens}
\end{figure}

\revise{
\subsection{Insight of PEFT in the Spectral Domain}
This section provides an intuitive understanding of the differences between spectral and spatial tuning and the insight of PEFT in the spectral domain.

The spatial domain tuning methods utilize tuning modules to process tokens representing localized patches of the point cloud, each corresponding to a specific region of points. As illustrated in Fig.~\ref{fig:spatial_and_spectral_tokens}(a), the zoomed-in area denotes one of these spatial tokens within a pre-trained model, where spatial fine-tuning adjusts the feature response at that patch. Besides, it can be found that even with similar geometric structures, features from the two plane wings of the point cloud show significant differences in output (reflected by their low cosine similarity). We argue that the pre-training tasks prioritize general point cloud representations without prior downstream tasks, resulting in the frozen model's difficulty in distinguishing fine-grained point cloud structures in the spatial domain.

Our PointGST proposes the spectral domain tuning paradigm. We transfer the point tokens from the spatial domain into the spectral domain via the Graph Fourier Transform (GFT) theory. A toy example is shown in Fig.~\ref{fig:spatial_and_spectral_tokens}(b), we can decompose the original features $\boldsymbol{T} \in \mathbb{R}^{n \times r}$ into $n$ orthogonal Laplacian eigenvectors, where each eigenvector corresponds to a graph signal with a distinct level of smoothness. It is worth noting that the Laplace matrix is constructed based on the geometric attributes of each point cloud instance, and the resulting eigenvectors are shared across different Transformer layers, thus involving intrinsic information about the downstream point clouds. Then, the goal of spectral domain fine-tuning is to modify the projected coefficients or the position ($\cdots,\boldsymbol{T}_{i}^{f},\boldsymbol{T}_{j}^{f},\boldsymbol{T}_{k}^{f},\cdots$) of $\boldsymbol{T}$ within the $n$-dimensional eigenvector space. 

In our context, the graph signal (compressed point tokens) $\boldsymbol{T}$ is defined on $n$ vertices. For simplicity, we assume $r=1$ here without loss of generality, as cases with $r>1$ can be reduced to multiple $r=1$ scenarios. Thus, Eq.~\ref{eq:gft-method} and Eq.~\ref{eq:igft-method} for $\boldsymbol{T} \in \mathbb{R}^{n \times 1}$ can be rewritten as:
\begin{equation}
\boldsymbol{T}^{f}_{i}=\sum_{k=0}^{n-1}\boldsymbol{U}_{k,i} \boldsymbol{T}_{k} = \langle\mathbf{u}_i, \boldsymbol{T}\rangle,
~~\boldsymbol{T}=\sum_{i=0}^{n-1}\boldsymbol{T}^f_{i}\mathbf{u}_{i},
\end{equation}
where ${\boldsymbol{T}}^f_{i} \in \mathbb{R}$ indicates the projection of $\boldsymbol{T}$ onto the $i$-th graph Fourier basis vector $\mathbf{u}_{i}$ and also the $i$-th Fourier coefficient, as shown in Fig.~\ref{fig:spatial_and_spectral_tokens}(b). In the GFT, the Laplace matrix $\boldsymbol{L}$ functions similarly to the Laplace operator in the Fourier transform, using the eigenvectors of $\boldsymbol{L}$ as the basis for graph Fourier projection. $\boldsymbol{L}$ is defined as the difference between the degree matrix and the adjacency matrix, measuring variations of the graph through edge-based differentiation. We have the total variation defined as below:
\begin{equation}
\begin{aligned}\label{eq:s2def}
\boldsymbol{T}^{\top}\boldsymbol{L} \boldsymbol{T}
= {\boldsymbol{T}^f}^\top \boldsymbol{\Lambda} \boldsymbol{T}^f
= \sum_{i=0}^{n-1} \lambda_i {\boldsymbol{T}^f_i}^2.
\end{aligned}
\end{equation}
From Eq.~\ref{eq:s2def}, it can be proved that to select a set of orthogonal graph signals with minimized total variances, those signals must be the eigenvectors of the Laplacian matrix~\cite{shuman2013emerging}. Consequently, $\mathbf{u}_{0}$ corresponds to the smoothest graph signal and $\mathbf{u}_{n-1}$ to the least smooth, as they relate to the smallest eigenvalues $\lambda_{0}$ and the largest $\lambda_{n-1}$, respectively. Thus, the GFT decomposes a graph signal into components of varying smoothness, analogous to how the Fourier transform separates a function into different frequencies.

To summarize, intuitively, the GFT decomposes a graph signal (in our case, point tokens) into orthogonal graph signals of varying smoothness (eigenvectors of the Laplace matrix), enabling the original point token to be expressed as a linear combination of these signals. Using eigenvectors as a basis helps alleviate spatial domain confusion in the spectral domain.
} 

\begin{table*}[htbp]
    \scriptsize
    \setlength{\tabcolsep}{2.4mm}
    \centering
  \caption{ The comparison between other fine-tuning strategies and our PointGST. We report the overall accuracy (OA) and trainable parameters (Params.) across three variants of ScanObjectNN~\cite{uy2019revisiting} and ModelNet40~\cite{wu20153d}. All methods use default data augmentation as the baseline. ScanObjectNN results are reported without voting, while ModelNet40 results are presented both with and without voting (-/-).
}
    \label{tab:different_finetune}
    \begin{tabular}{lcccccccccc}
    \toprule
    \multirow{2.3}{*}{Pre-trained model} &\multirow{2.3}{*}{Fine-tuning strategy} &\multirow{2.3}{*}{Reference} &\multirow{2.3}{*}{Params. (M)} &\multirow{2.3}{*}{FLOPs (G)} &\multicolumn{3}{c}{ScanObjectNN} & ModelNet40\\
    \cmidrule(lr){6-9}
    & & & & &OBJ\_BG & OBJ\_ONLY &PB\_T50\_RS & OA (\%)      \\
    \midrule
    \multirow{5}{*}{\tabincell{c}{Point-BERT~\cite{yu2022point}\\(CVPR 22)}} & \textcolor{gray}{Fully fine-tune} & \textcolor{gray}{-} & \textcolor{gray}{22.1 (100\%)}  & \textcolor{gray}{4.76} & \textcolor{gray}{87.43} & \textcolor{gray}{88.12} & \textcolor{gray}{83.07} & \textcolor{gray}{92.7} / {\color{gray}{93.2}}\\
    & IDPT~\cite{zha2023instance}& ICCV 23 & 1.7 (7.69\%) & 7.10 & 88.12\dplus{+0.69} & 88.30\dplus{+0.18} & 83.69\dplus{+0.62}   & 92.6{\dtplus{-0.1}} / {\color{gray}{93.4}}\dplus{+0.2}\\
    & Point-PEFT~\cite{tang2024point}& AAAI 24 & 0.7 (3.13\%)  & 7.61 & 88.81\dplus{+1.38} & 89.67\dplus{+1.55} & 85.00\dplus{+1.93} & {93.4{\dplus{+0.7}} / ~~~-~~~~~~~~} \\
    & DAPT~\cite{zhou2024dynamic}& CVPR 24 & 1.1 (4.97\%)  & 4.96 & 91.05\dplus{+3.62} & 89.67\dplus{+1.55} & 85.43\dplus{+2.36}  & 93.1{\dplus{+0.4}} / {\color{gray}{93.6}}\dplus{+0.4} \\
    & PointGST (\textbf{ours})& -  & \textbf{0.6 (2.77\%)}  & \textbf{4.81} & \textbf{91.39}\dplus{+3.96} & \textbf{89.67}\dplus{+1.55} & \textbf{85.64}\dplus{+2.57}  & \textbf{93.4}{\dplus{+0.7}} / {\color{gray}{93.8}}\dplus{+0.6}\\
    \midrule
    
    \multirow{5}{*}{\tabincell{c}{Point-MAE~\cite{pang2022masked}\\(ECCV 22)}}  & \textcolor{gray}{Fully fine-tune} & \textcolor{gray}{-}& \textcolor{gray}{22.1 (100\%)}  & \textcolor{gray}{4.76} & \textcolor{gray}{90.02} & \textcolor{gray}{88.29} & \textcolor{gray}{85.18}  & \textcolor{gray}{93.2} / {\color{gray}{93.8}}\\
    &IDPT~\cite{zha2023instance}& ICCV 23 & 1.7 (7.69\%)  & 7.10 & 91.22\dplus{+1.20} & 90.02\dplus{+1.73} & 84.94\dtplus{-0.24}  & 93.3{\dplus{+0.1}} / {\color{gray}{94.4}}\dplus{+0.6} \\
    & Point-PEFT~\cite{tang2024point} & AAAI 24 & 0.7 (3.13\%)  & 7.61 & 89.67\dtplus{-0.35} & 88.98\dplus{+0.69} & 84.91\dtplus{-0.27}  & 93.3{\dplus{+0.1}} / {\color{gray}{93.9}\dplus{+0.1}}\\
    &DAPT~\cite{zhou2024dynamic}& CVPR 24 & 1.1 (4.97\%) & 4.96 & 90.88\dplus{+0.86} & 90.19\dplus{+1.90} & 85.08\dtplus{-0.10}  & 93.5{\dplus{+0.3}} / {\color{gray}{{94.0}}}\dplus{+0.2} \\
    & PointGST (\textbf{ours})& -  & \textbf{0.6 (2.77\%)}  & \textbf{4.81} & \textbf{91.74}\dplus{+1.72} & \textbf{90.19}\dplus{+1.90} & \textbf{85.29}\dplus{+0.11}  & \textbf{93.5}{\dplus{+0.3}} / {\color{gray}{94.0}}\dplus{+0.2}\\
    \midrule
    
    \multirow{5}{*}{\tabincell{c}{~~~~~ACT~\cite{dong2023act}\\~~~~(ICLR 23)}}& \textcolor{gray}{Fully fine-tune} & \textcolor{gray}{-} & \textcolor{gray}{22.1 (100\%)} & \textcolor{gray}{4.76} & \textcolor{gray}{93.29} & \textcolor{gray}{91.91} & \textcolor{gray}{88.21}  & ~\textcolor{gray}{-}~~~ / {\color{gray}{93.7}}~~~~~\\
    & IDPT~\cite{zha2023instance}& ICCV 23 & 1.7 (7.69\%)  & 7.10 & 93.12\dtplus{-0.17} & 92.26\dplus{+0.35} & 87.65\dtplus{-0.56}  & ~~~-~~~ / {\color{gray}{94.0}}\dplus{+0.3} \\
    & Point-PEFT~\cite{tang2024point}& AAAI 24 & 0.7 (3.13\%)  & 7.61 & 90.36\dtplus{-2.93} & 90.02\dtplus{-1.89} & 85.74\dtplus{-2.47} & 93.1 / {\color{gray}{93.4}}\dtplus{-0.3}\\
    &  DAPT~\cite{zhou2024dynamic}& CVPR 24 & 1.1 (4.97\%) & 4.96 & 92.60\dtplus{-0.69} & 91.57\dtplus{-0.34} & 87.54\dtplus{-0.67}  & 92.7 / {\color{gray}{93.2}}\dtplus{-0.5} \\
    &  PointGST (\textbf{ours})& -  & \textbf{0.6 (2.77\%)}  & \textbf{4.81} & \textbf{93.46}\dplus{+0.17} & \textbf{92.60}\dplus{+0.69} & \textbf{88.27}\dplus{+0.06}  & \textbf{93.4} / {\color{gray}{94.0}}\dplus{+0.3}\\
    \midrule
    
    \multirow{5}{*}{\tabincell{c}{~~~\recon~\cite{qi2023contrast}\\~~(ICML 23)}}& \textcolor{gray}{Fully fine-tune} & \textcolor{gray}{-} & \textcolor{gray}{22.1 (100\%)}  & \textcolor{gray}{4.76} & \textcolor{gray}{94.32} & \textcolor{gray}{92.77} & \textcolor{gray}{90.01} & \textcolor{gray}{92.5} / {\color{gray}{93.0}}\\
    & IDPT~\cite{zha2023instance}& ICCV 23  &1.7 (7.69\%)  & 7.10 &93.29\dtplus{-1.03} &91.57\dtplus{-1.20} & 87.27\dtplus{-2.74}  & 93.4{\dplus{+0.9}} / {\color{gray}{{93.5}}}\dplus{+0.5} \\ 
    & Point-PEFT~\cite{tang2024point}& AAAI 24 & 0.7 (3.13\%)  & 7.61 & 91.91\dtplus{-2.41} & 90.19\dtplus{-2.58} & 86.36\dtplus{-3.65} & 93.3{\dplus{+0.8}} / {\color{gray}{93.8}}\dplus{+0.8}\\ 
    & DAPT~\cite{zhou2024dynamic}& CVPR 24  & 1.1 (4.97\%)  & 4.96 &94.32\dtplus{-0.00} & 92.43\dtplus{-0.34} & 89.38\dtplus{-0.63}  & 93.5{\dplus{+1.0}} / {\color{gray}{94.1}}\dplus{+1.1}\\
    & PointGST (\textbf{ours})& -  & \textbf{0.6 (2.77\%)}  & \textbf{4.81} & \textbf{94.49}\dplus{+0.17} & \textbf{92.94}\dplus{+0.17} & \textbf{89.49}\dtplus{-0.52}  & \textbf{93.6}{\dplus{+1.1}} / {\color{gray}{94.1}}\dplus{+1.1}\\
    \midrule
    
    \multirow{5}{*}{\tabincell{c}{~~PointGPT-L~\cite{chen2024pointgpt}\\~(NeurIPS 23)}} & \textcolor{gray}{Fully fine-tune} & \textcolor{gray}{-} & \textcolor{gray}{360.5 (100\%)}  & \textcolor{gray}{67.71} & \textcolor{gray}{97.2} & \textcolor{gray}{96.6} & \textcolor{gray}{93.4} & \textcolor{gray}{94.1} / {\color{gray}{94.7}}\\
    & IDPT~\cite{zha2023instance}& ICCV 23 & 10.0 (2.77\%) & 75.19 & 98.11\dplus{+0.91} & 96.04\dtplus{-0.56} & 92.99\dtplus{-0.41}   & 93.4{\dtplus{-0.7}} / {\color{gray}{94.6}}\dtplus{-0.1}\\
    & Point-PEFT~\cite{tang2024point} & AAAI 24 & 3.1 (0.86\%)  & 73.62 & 96.39\dtplus{-0.81} & 94.66\dtplus{-1.94} & 92.85\dtplus{-0.55} & 93.5{\dtplus{-0.6}} / {\color{gray}{94.2}}\dtplus{-0.5}\\
    & DAPT~\cite{zhou2024dynamic}& CVPR 24 & 4.2 (1.17\%) & 71.64 & 98.11\dplus{+0.91} & 96.21\dtplus{-0.39} & 93.02\dtplus{-0.38}  & 94.2{\dplus{+0.1}} / {\color{gray}{94.9}}\dplus{+0.2} \\
    & PointGST (\textbf{ours})& -  & \textbf{2.4 (0.67\%)}  & \textbf{67.95} & \textbf{98.97}\dplus{+1.77} & \textbf{97.59}\dplus{+0.99} & \textbf{94.83}\dplus{+1.43}   &  \textbf{94.8}{\dplus{+0.7}} / {\color{gray}{95.3}}\dplus{+0.6}\\
    \bottomrule
    \end{tabular}
\end{table*}

\begin{table}[!t]
\scriptsize
\setlength{\tabcolsep}{1mm}
\centering
\caption{Classification on ScanObjectNN PB\_T50\_RS~\cite{uy2019revisiting} with strong data augmentation. Overall accuracy (\%) without voting is reported. Params. is the trainable parameters. }
\label{tab:wDataArgumentaion}
\begin{tabular}{ lcccc }
\toprule
    Pre-trained model & Fine-tuning strategy  &Reference& Params. (M) & PB\_T50\_RS \\
 \midrule
     \multirow{4}{*}{\tabincell{c}{Point-MAE~\cite{pang2022masked}\\(ECCV 22)}} & \textcolor{gray}{Fully fine-tune} & \textcolor{gray}{-} & \textcolor{gray}{22.1} & \textcolor{gray}{88.4} \\
 & DA~\cite{fei2024fine}& ICRA 24 & 1.6 &88.0\\
 & Point-PEFT~\cite{tang2024point}&AAAI 24 &0.7 & 89.1\\
& PointGST (\textbf{ours}) & - & \textbf{0.6} & \textbf{89.3} \\
\bottomrule
\end{tabular}
\end{table}

\section{EXPERIMENTAL Setup}

This section describes the implementation details and the comparison methods. 

\subsection{Datasets and Implementation Details}
All the experiments are conducted on a single 24 GB NVIDIA GPU. During fine-tuning, the parameters of the pre-trained backbones remain frozen, while only the newly added parameters are updated. \revise{The hyper-parameters, scaling factor $s$, group number $k$, and adapter dimension $r$ are set to 1, 4, and 36, respectively. For all conducted baselines, these hyper-parameters are consistently applied across all datasets/tasks without specific adjustments.} We employ a transpose format of Z space-filling curve~\cite{morton1966computer} to scan the key points to construct the local graphs.

\textbf{ScanObjectNN}~\cite{uy2019revisiting} is a challenging real-world 3D object dataset with $\sim$15K indoor point cloud instances across 15 categories, including three variants (OBJ\_BG, OBJ\_ONLY, PB\_T50\_RS) of increasing difficulty. We follow baseline training settings: AdamW optimizer~\cite{loshchilov2019decoupled} (weight decay 0.05), cosine learning rate schedule~\cite{loshchilov2017sgdr} (initial LR 5e-4, 10-epoch warm-up), 300 epochs, batch size 32, and 2048 input points (divided into 128 groups of 32 points each).

\textbf{ModelNet40}~\cite{wu20153d} is a synthetic dataset of 12,311 CAD models from 40 categories. The training follows ScanObjectNN settings, with 1024 input points (divided into 64 groups of 32 points each) and 300 epochs.

\textbf{ShapeNetPart}~\cite{yi2016scalable} is a point-level part segmentation benchmark with 16,881 instances from 16 objects and 50 part categories. We use AdamW (weight decay 1e-4), cosine scheduler (initial LR 1e-4, 10-epoch warm-up), 300 epochs, batch size 16, and 2048 points (divided into 128 groups of 32 points each).

\textbf{S3DIS}~\cite{armeni20163d} contains six large-scale indoor areas with 273M points over 13 categories. Following~\cite{tchapmi2017segcloud}, Area 5 is used for evaluation. Models are trained for 60 epochs (batch size 32, initial LR 2e-4, cosine scheduler). Other settings follow ShapeNetPart.

\textbf{ScanNetV2}~\cite{dai2017scannet} comprises 1,201 training, 312 validation, and 100 test scenes with 18 annotated categories. We use AdamW (weight decay 0.1), cosine scheduler (initial LR 2e-4, 10-epoch warm-up), 1080 epochs, and batch size 8.

\Revise{\textbf{PCN}~\cite{yuan2018pcn} contains 30,974 pairs of point clouds from 8 categories. The network takes 2048 points as inputs for each and completes the other 14,336 points. We use AdamW (weight decay 5e-4), cosine scheduler (initial LR 1e-4), 100 epochs, and batch size 16.}

\subsection{ Baseline Methods and Comparisons}
To illustrate the effectiveness of our approach, popular pre-trained models like Point-BERT~\cite{yu2022point}, Point-MAE~\cite{pang2022masked}, ACT~\cite{dong2023act}, \recon~\cite{qi2023contrast}, and PointGPT~\cite{chen2024pointgpt} are selected as our baselines \revise{on all datasets}. Additionally, we include the SOTA point cloud PEFT methods for comparisons: 1) \textbf{IDPT}~\cite{zha2023instance} utilizes DGCNN~\cite{wang2019dynamic} to generate instance-aware prompts before the last transformer layer for model fine-tuning rather than relying on static prompts; 2) \textbf{Point-PEFT}~\cite{tang2024point} utilizes a point-prior bank for feature aggregation and incorporating Adapters into each transformer block. 3) \textbf{DA}~\cite{fei2024fine} uses a dynamic aggregation method to replace previous static aggregation for pre-trained point cloud Transformers. 4) \textbf{DAPT}~\cite{zhou2024dynamic}, the latest SOTA point cloud PEFT method, proposes to adjust tokens based on significance scores. It should be noted that for datasets or tasks where certain methods have not provided results, we reproduce their performance using the official code to ensure a comprehensive comparison.

\section{Results and Analysis}
In this section, we conduct experiments to show the effectiveness of our approach. 

\subsection{3D Classification}

\subsubsection{The comparisons of different fine-tuning strategies}
\label{sec:comparison_dapt_idpt}

We first comprehensively compare the representative fine-tuning strategies used in the point cloud tasks, including the fully fine-tuning (FFT), IDPT~\cite{zha2023instance}, Point-PEFT~\cite{tang2024point}, and DAPT~\cite{zhou2024dynamic}, in terms of trainable parameters (Params.) and performance. We apply our PointGST to various pre-trained models~\cite{yu2022point, pang2022masked, dong2023act, qi2023contrast, chen2024pointgpt}, size ranging from 22.1M to 360.5M. Tab.~\ref{tab:different_finetune} lists the detailed results, and we found that:

\begin{table*}[ht]
    \scriptsize
    \setlength{\tabcolsep}{5.mm}
    \centering
  \caption{
Comparison of overall accuracy (OA) and trainable parameters (Params.) across three variants of ScanObjectNN~\cite{uy2019revisiting} and ModelNet40~\cite{wu20153d}. Results highlighted in \colorbox{greenbg}{green} indicate the use of the voting strategy.
}
    \label{tab:compare_sota}
    \begin{tabular}{lccccccc}
    \toprule
    \multirow{2.3}{*}{Method} &\multirow{2.3}{*}{Reference} &\multirow{2.3}{*}{Params. (M)} &\multicolumn{3}{c}{ScanObjectNN} &\multicolumn{2}{c}{ModelNet40}\\
		\cmidrule(lr){4-6} \cmidrule(lr){7-8}
	& & & OBJ\_BG & OBJ\_ONLY &PB\_T50\_RS & Points Num. & OA (\%)      \\
    \midrule
    \multicolumn{8}{c}{\textit{Supervised Learning Only}} \\
    \midrule
    PointNet~\cite{qi2017pointnet} & CVPR 17 & 3.5  & 73.3  & 79.2  & 68.0 & 1k & \cellcolor{greenbg}89.2 \\
    PointNet++~\cite{qi2017pointnet++}   & NeurIPS 17 & 1.5  & 82.3  & 84.3  & 77.9 & 1k & \cellcolor{greenbg}90.7\\
    DGCNN~\cite{wang2019dynamic}  & TOG 19 & 1.8  & 82.8  & 86.2  & 78.1 & 1k & \cellcolor{greenbg}92.9 \\
    MVTN~\cite{hamdi2021mvtn}  & ICCV 21 & 11.2  & -     & -     & 82.8 & 1k & \cellcolor{greenbg}93.8\\
    3D-GCN~\cite{lin2021learning} & TPAMI 22& -  & - & - & - & 1k & \cellcolor{greenbg}92.1\\
    RepSurf-U~\cite{ran2022surface} & CVPR 22 & 1.5  &  -  & -    & 84.3  & 1k  & \cellcolor{greenbg}94.4 \\
    PointNeXt~\cite{qian2022pointnext}  & NeurIPS 22  & 1.4  & -     & -  & 87.7 & 1k & \cellcolor{greenbg}94.0\\
    PTv2~\cite{wu2022point}  & NeurIPS 22  & 12.8  & - & - & -  & 1k & \cellcolor{greenbg}94.2\\
    PointMLP~\cite{ma2022rethinking}  & ICLR 22 &  13.2  & -    & -     & 85.4  & 1k & \cellcolor{greenbg}94.5\\
    PointMeta~\cite{lin2023meta} & CVPR 23 & -  &  - & -   & 87.9 & -  & \cellcolor{greenbg}- \\
    ADS~\cite{hong2023attention} & ICCV 23 & -  &  - & -   & 87.5 & 1k  & \cellcolor{greenbg}95.1 \\
    X-3D~\cite{sun2024x} & CVPR 24 & 5.4  &  - & -   & 90.7 & -  & \cellcolor{greenbg}- \\
    GPSFormer~\cite{wang2024gpsformer} & ECCV 24 & 2.4  &  - & -   & 95.4\cellcolor{greenbg} & 1k  & \cellcolor{greenbg}94.2 \\
    \midrule
    \multicolumn{8}{c}{\textit{ Self-Supervised Representation Learning (Full fine-tuning)}} \\
    \midrule
    OcCo~\cite{wang2021unsupervised} & ICCV 21 & 22.1  & 84.85 & 85.54 & 78.79 & 1k & \cellcolor{greenbg}92.1 \\
    Point-BERT~\cite{yu2022point}  & CVPR 22 & 22.1  & 87.43 & 88.12 &  83.07 & 1k & \cellcolor{greenbg}93.2 \\
    MaskPoint~\cite{liu2022masked} & ECCV 22 & 22.1  & 89.70 & 89.30 &  84.60 & 1k & \cellcolor{greenbg}93.8 \\
    Point-MAE~\cite{pang2022masked}  & ECCV 22 & 22.1  & 90.02 & 88.29 & 85.18 & 1k & \cellcolor{greenbg}93.8 \\
    Point-M2AE~\cite{zhang2022point}  & NeurIPS 22 & 15.3  & 91.22 & 88.81 & 86.43  & 1k & \cellcolor{greenbg}94.0\\
    ACT~\cite{dong2023act}   & ICLR 23 & 22.1  & 93.29 & 91.91  & 88.21 & 1k & \cellcolor{greenbg}93.7\\
    \recon~\cite{qi2023contrast} & ICML 23& 43.6  & 94.15 & 93.12  & 89.73 & 1k & \cellcolor{greenbg}93.9  \\
    PointGPT-L~\cite{chen2024pointgpt}  & NeurIPS 23 & 360.5  & 97.2 & 96.6 & 93.4& 1k & \cellcolor{greenbg}94.7\\
    Point-FEMAE~\cite{zha2024towards}  & AAAI 24 & 27.4  & 95.18 & 93.29 & 90.22 & 1k &  \cellcolor{greenbg}94.5\\
    P2P++~\cite{wang2024point}  & TPAMI 24 & 16.1  & - & - & 90.3 & 1k &  \cellcolor{greenbg}94.1\\
    PointDif~\cite{zheng2024point}  & CVPR 24 & -  & 93.29 & 91.91 & 87.61 & - & \cellcolor{greenbg}- \\
    PointMamba~\cite{liang2024pointmamba}&NeurIPS 24& 12.3  & 94.32 & 92.60 & 89.31 & - &  \cellcolor{greenbg}-\\
    MH-PH~\cite{feng2024shape2scene}  & ECCV 24 & -  & 97.4 & 96.8 & 93.8& 1k &  \cellcolor{greenbg}94.6\\
    \recon++-L\textcolor{red}~\cite{qi2024shapellm} &ECCV 24 & 657.2  & \cellcolor{greenbg}98.80 & \cellcolor{greenbg}97.59 & \cellcolor{greenbg}95.25 & 1k &\cellcolor{greenbg}94.8\\
    \midrule
    \multicolumn{8}{c}{\textit{Self-Supervised Representation Learning (Efficient fine-tuning)}} \\
    \midrule
    PointGST (\textbf{ours})& -  & \textbf{2.4}  & 98.97 & 97.59 & 94.83& 1k & 94.8 \\
    PointGST (\textbf{ours})& -  & \textbf{2.4} & \cellcolor{greenbg}\textbf{99.48}  & \cellcolor{greenbg}\textbf{97.76} & \cellcolor{greenbg}\textbf{96.18} &1k & \cellcolor{greenbg}\textbf{95.3}\\
    \bottomrule
    \end{tabular}%
\end{table*}

\textbf{1) PointGST effectively balances the number of trainable parameters and performance.} 
Based on the quantitative comparisons, it is evident that increasing the scale of a pre-trained model will improve the performance. However, this would meanwhile come at a huge cost of fine-tuning, e.g., Point-MAE and PointGPT-L report 85.18\% and 93.4\% with 22.1M and 360.5M parameters on the ScanObjectNN PB\_T50\_RS dataset, respectively. Thanks to only fine-tuning a few parameters in the spectral domain, our method can effectively address this dilemma. Specifically, on the most convincing pre-trained point cloud models, PointGPT-L~\cite{chen2024pointgpt}, compared with the fully fine-tuning (FFT), we only need 2.4M trainable parameters, which is merely 0.67\% of the FFT. More importantly, with so few training parameters, we even outperform the FFT by 1.77\%, 0.99\%, 1.43\%, and 0.7\% on the OBJ\_BG, OBJ\_ONLY, PB\_T50\_RS and ModelNet40 datasets, respectively. Compared to spatial domain-based methods like IDPT, Point-PEFT, and DAPT, our PointGST achieves superior performance with fewer trainable parameters by leveraging the spectral domain, where features are more easily de-correlated. In short, PointGST proves to be a potential candidate that brings promising performance and parameter-efﬁcient approaches to low-resource scenarios.

\textbf{2) PointGST can effectively generalize to diverse pre-trained models.} A good PEFT method should consistently perform well across different pre-trained models, regardless of the pre-training strategies, data used, or model sizes. However, our observations reveal that existing PEFT methods~\cite{zha2023instance,zhou2024dynamic,tang2024point,fei2024fine} fail to deliver consistent improvements across diverse pre-trained models. For instance, as shown in Tab.~\ref{tab:different_finetune}, we apply five distinct pre-trained models, varying in pre-training techniques (e.g., masked modeling and contrastive learning) and sizes (ranging from 22.1M to 360.5M parameters). Notably, IDPT, Point-PEFT, and DAPT exhibit negative impacts when applied to baselines like ACT~\cite{dong2023act} and \recon~\cite{qi2023contrast}. Even in scenarios where these methods produce positive outcomes, the gains from IDPT, Point-PEFT, and DAPT remain limited. In contrast, our PointGST consistently yields improvements across most conducted pre-trained models and outperforms existing point cloud PEFT methods. In addition, we find that both Point-PEFT~\cite{tang2024point} and DA~\cite{fei2024fine} apply the point cloud rotation as the data augmentation~\cite{dong2023act} on the Point-MAE. To keep fair, we also adopt the augmentations to conduct experiments. The comparisons in Tab.~\ref{tab:wDataArgumentaion} clearly show that PointGST consistently achieves higher accuracy than Point-PEFT and DA.

\begin{table}[!t]
  \centering
  \scriptsize
    \setlength{\tabcolsep}{0.5mm}
  \caption{Few-shot learning on ModelNet40~\cite{wu20153d}. Overall accuracy$\pm$standard deviation without voting is reported.}
    \begin{tabular}{lccccc}
    \toprule
   \multirow{2.3}{*}{Methods}&\multirow{2.3}{*}{Reference} & \multicolumn{2}{c}{5-way} & \multicolumn{2}{c}{10-way} \\
\cmidrule(lr){3-4}\cmidrule(lr){5-6}  &        & 10-shot & 20-shot & 10-shot & 20-shot \\
    \midrule
   \textcolor{gray}{Point-BERT~\cite{yu2022point} (baseline)}  &\textcolor{gray}{CVPR 22} &\textcolor{gray}{94.6$\pm$3.1} & \textcolor{gray}{96.3$\pm$2.7} & \textcolor{gray}{91.0$\pm$5.4} & \textcolor{gray}{92.7$\pm$5.1} \\
   + IDPT~\cite{zha2023instance}  & ICCV 23    & 96.0$\pm$1.7& 97.2$\pm$2.6& 91.9$\pm$4.4& 93.6$\pm$3.5\\
   + Point-PEFT~\cite{tang2024point}  & AAAI 24 & 95.4$\pm$3.0 & 97.3$\pm$1.9 & 91.6$\pm$4.5 & 94.5$\pm$3.5 \\
   + DAPT~\cite{zhou2024dynamic} & CVPR 24 &95.8$\pm$2.1 &97.3$\pm$1.3&92.2$\pm$4.3&94.2$\pm$3.4 \\
   + PointGST (\textbf{ours}) & -&\textbf{96.5}$\pm$2.4 &\textbf{97.9}$\pm$2.0 &\textbf{92.7}$\pm$4.2 &\textbf{95.0}$\pm$2.8 \\
    \midrule
    \textcolor{gray}{Point-MAE~~\cite{pang2022masked} (baseline)}&\textcolor{gray}{ECCV 22} & \textcolor{gray}{96.3$\pm$2.5} & \textcolor{gray}{97.8$\pm$1.8} & \textcolor{gray}{92.6$\pm$4.1} & \textcolor{gray}{95.0$\pm$3.0}\\
   + IDPT~\cite{zha2023instance} &   ICCV 23    & 97.3$\pm$2.1& 97.9$\pm$1.1& 92.8$\pm$4.1& 95.4$\pm$2.9\\
   + Point-PEFT~\cite{tang2024point}  & AAAI 24 & 95.5$\pm$2.9 & 97.6$\pm$1.7 & 91.7$\pm$4.3 & 94.7$\pm$3.0 \\
   + DAPT~\cite{zhou2024dynamic} & CVPR 24 & 96.8$\pm$1.8  & 98.0$\pm$1.0 & 93.0$\pm$3.5 & 95.5$\pm$3.2  \\
   + PointGST (\textbf{ours}) & -&\textbf{98.0}$\pm$1.8 &\textbf{98.3}$\pm$0.9 &\textbf{93.7}$\pm$4.0 &\textbf{95.7}$\pm$2.4 \\
    \midrule
    \revise{\textcolor{gray}{ACT~~\cite{dong2023act} (baseline)}} & \revise{\textcolor{gray}{ICLR 23}} & \revise{\textcolor{gray}{96.8$\pm$2.3}} & \revise{\textcolor{gray}{98.0$\pm$1.4}} & \revise{\textcolor{gray}{93.3$\pm$4.0}} & \revise{\textcolor{gray}{95.6$\pm$2.8}} \\
    \revise{+ IDPT~\cite{zha2023instance}} & \revise{ICCV 23} & \revise{96.7$\pm$2.6} & \revise{98.2$\pm$0.9} & \revise{92.4$\pm$4.5} & \revise{95.5$\pm$3.0} \\
    \revise{+ Point-PEFT}~\cite{tang2024point}  &\revise{ AAAI 24 }&\revise{ 95.4$\pm$2.7 }& \revise{97.7$\pm$1.7 }& \revise{91.8$\pm$5.1 }& \revise{94.9$\pm$3.0} \\
    \revise{+ DAPT}~\cite{zhou2024dynamic} & \revise{CVPR 24} & \revise{94.4$\pm$3.9} & \revise{97.3$\pm$2.4} & \revise{90.6$\pm$6.1} & \revise{95.1$\pm$3.7} \\
    \revise{+ PointGST (\textbf{ours})} & \revise{-} & \revise{\textbf{97.2}$\pm$1.9} & \revise{\textbf{98.6}$\pm$1.5} & \revise{\textbf{92.8}$\pm$4.0} & \revise{\textbf{95.5}$\pm$3.1} \\
    \midrule
    \revise{\textcolor{gray}{\recon~~\cite{qi2023contrast} (baseline)}} & \revise{\textcolor{gray}{ICML 23}} & \revise{\textcolor{gray}{97.3$\pm$1.9}} & \revise{\textcolor{gray}{98.9$\pm$1.2}} & \revise{\textcolor{gray}{93.3$\pm$3.9}} & \revise{\textcolor{gray}{95.8$\pm$3.0}} \\
    \revise{+ IDPT}~\cite{zha2023instance} & \revise{ICCV 23} & \revise{96.8$\pm$2.2} & \revise{98.6$\pm$0.8} & \revise{92.7$\pm$3.8} & \revise{\textbf{95.9}$\pm$3.2} \\
    \revise{+ Point-PEFT}~\cite{tang2024point}& \revise{AAAI 24 }& \revise{95.4$\pm$2.6} & \revise{97.7$\pm$1.4} & \revise{91.5$\pm$5.0} & \revise{95.2$\pm$3.4} \\
    \revise{+ DAPT}~\cite{zhou2024dynamic} & \revise{CVPR 24} & \revise{95.6$\pm$3.4} & \revise{97.3$\pm$2.0} & \revise{91.9$\pm$4.9} & \revise{94.5$\pm$3.2} \\
    \revise{+ PointGST (\textbf{ours})} & \revise{-} & \revise{\textbf{96.9}$\pm$2.2} & \revise{\textbf{98.7}$\pm$0.9} & \revise{\textbf{92.9}$\pm$3.8} & \revise{95.8$\pm$2.8} \\
    \midrule
    \revise{\textcolor{gray}{PointGPT-L~~\cite{chen2024pointgpt} (baseline)}} & \revise{\textcolor{gray}{NeurIPS 23}} & \revise{\textcolor{gray}{98.0$\pm$1.9}} & \revise{\textcolor{gray}{99.0$\pm$1.0}} & \revise{\textcolor{gray}{94.1$\pm$3.3}} & \revise{\textcolor{gray}{96.1$\pm$2.8}} \\
    \revise{+ IDPT}~\cite{zha2023instance} & \revise{ICCV 23} & \revise{96.8$\pm$2.0} & \revise{98.3$\pm$1.4} & \revise{95.3$\pm$2.5} & \revise{96.5$\pm$2.5} \\
    \revise{+ Point- 
    PEFT}~\cite{tang2024point}  & \revise{AAAI 24} & \revise{95.7$\pm$2.5} & \revise{98.0$\pm$1.8} & \revise{94.8$\pm$2.7} & \revise{96.3$\pm$2.4} \\
    \revise{+ DAPT}~\cite{zhou2024dynamic} & \revise{CVPR 24} & \revise{97.2$\pm$3.4} & \revise{98.4$\pm$1.3} & \revise{95.9$\pm$2.7} & \revise{97.3$\pm$2.4} \\
    \revise{+ PointGST (\textbf{ours})} & \revise{-} & \revise{\textbf{97.4}$\pm$2.0} & \revise{\textbf{99.6}$\pm$0.5} & \revise{\textbf{96.3}$\pm$3.6} & \revise{\textbf{97.6}$\pm$2.9} \\
    \bottomrule
    \end{tabular}
  
  \label{tab:fewshot}
\end{table}

\subsubsection{Compared with state-of-the-art methods}
The comparisons between the state-of-the-art (SOTA) methods and our proposed method on ScanObjectNN and ModelNet40 are illustrated in Tab.~\ref{tab:compare_sota}. We categorize the comparison methods into two types: supervised learning only and self-supervised representation learning with full fine-tuning. The results yield two notable observations regarding the efficacy of the proposed PointGST:

\textbf{1)} Self-supervised representation learning methods~\cite{chen2024pointgpt,qi2023contrast,dong2023act} generally outperform those based solely on supervised learning~\cite{ran2022surface,hong2023attention,ma2022rethinking}, emphasizing the critical role of pre-training. Additionally, these self-supervised methods often involve more trainable parameters than their supervised counterparts (e.g., 13.2M and 657.2M for PointMLP~\cite{ma2022rethinking} and \recon++-L~\cite{qi2024shapellm}, respectively), which demonstrates the practical relevance of parameter-efficient fine-tuning.

\textbf{2)} When using PointGPT-L~\cite{chen2024pointgpt} as the baseline, our PointGST outperforms all previous methods, establishing a new state-of-the-art, while requiring only 2.4M trainable parameters. To the best of our knowledge, this is the first approach to nearly saturate the performance on the ScanObjectNN OBJ\_BG dataset (e.g., 99.48\% OA). Note that this is not attributable to simply overfitting the dataset. Instead, it highlights the current limitations of existing point cloud analysis datasets in effectively evaluating new methods. Consequently, we encourage the community to develop more challenging datasets to better assess the progress in future point cloud analysis research.

\begin{figure}[t]
	\begin{center}
		\includegraphics[width=0.98\linewidth]{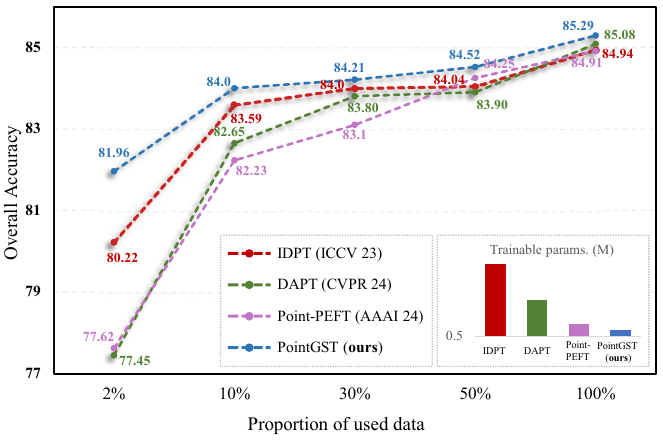}
	\end{center}
 \vspace{-12pt}
	\caption{Few-shot training performance on the ScanObjectNN PB\_T50\_RS dataset at different training data proportions.}
	\label{fig:few_show}
\end{figure}

\begin{table}[t]
  \centering
  \scriptsize
  \setlength{\tabcolsep}{0.2mm}
  \caption{Part segmentation on the ShapeNetPart~\cite{yi2016scalable}. The mIoU for all classes (Cls.) and for all instances (Inst.) are reported. Params. represents the trainable parameters. }
    \label{tab:segmentation}
    \begin{tabular}{lcccc}
    \toprule
    Methods & Reference & Params. (M)& Cls. mIoU (\%) & Inst. mIoU (\%) \\
    \midrule
    \textcolor{gray}{Point-BERT~\cite{yu2022point} (baseline)} &  \textcolor{gray}{CVPR 22} & \textcolor{gray}{27.06} & \textcolor{gray}{84.11} & \textcolor{gray}{85.6} \\ 
    + IDPT~\cite{zha2023instance} & ICCV 23 & 5.69  & 83.50  & 85.3  \\
    + Point-PEFT~\cite{tang2024point} & AAAI 24 & 5.62  & 81.12  &  84.3 \\
    + DAPT~\cite{zhou2024dynamic} & CVPR 24 & 5.65  & 83.83 & 85.5 \\
    + PointGST (\textbf{ours})& - & \textbf{5.58}  & \textbf{83.87} & \textbf{85.7} \\
    \midrule
    \textcolor{gray}{Point-MAE~\cite{pang2022masked} (baseline)} &  \textcolor{gray}{ECCV 22} & \textcolor{gray}{27.06} & \textcolor{gray}{84.19} & \textcolor{gray}{86.1} \\ 
    + IDPT~\cite{zha2023instance} & ICCV 23 & 5.69  & 83.79  & 85.7  \\
    + Point-PEFT~\cite{tang2024point} & AAAI 24 & 5.62  & 83.20 &  85.2 \\
    + DAPT~\cite{zhou2024dynamic} & CVPR 24 & 5.65  & \textbf{84.01} & 85.7 \\
    + PointGST (\textbf{ours})& - & \textbf{5.59}  & 83.81 & \textbf{85.8} \\
    \midrule
    \revise{\textcolor{gray}{ACT~\cite{dong2023act}~(baseline)}} & \revise{\textcolor{gray}{ICLR 23}} & \revise{\textcolor{gray}{27.06}} & \revise{\textcolor{gray}{84.66}} & \revise{\textcolor{gray}{86.1}} \\
    \revise{+ IDPT~\cite{zha2023instance}} & \revise{ICCV 23} & \revise{5.69} & \revise{83.82} & \revise{\textbf{85.9}} \\
    \revise{+ Point-PEFT~\cite{tang2024point}} & \revise{AAAI 24} & \revise{5.62} & \revise{81.25} & \revise{84.4} \\
    \revise{+ DAPT~\cite{zhou2024dynamic}} & \revise{CVPR 24} & \revise{5.65} & \revise{83.44} & \revise{85.5} \\
    \revise{+ PointGST (\textbf{ours})} & \revise{-} & \revise{\textbf{5.59}} & \revise{\textbf{84.04}} & \revise{85.8} \\
    \midrule
    \textcolor{gray}{\recon~\cite{qi2023contrast}~(baseline)} & \textcolor{gray}{ICML 23} & \textcolor{gray}{27.06}& \textcolor{gray}{84.52} & \textcolor{gray}{86.1} \\
    + IDPT~\cite{zha2023instance} &  ICCV 23 & 5.69 & 83.66  & 85.7 \\
    + Point-PEFT~\cite{tang2024point} & AAAI 24 & 5.62  & 83.10  &  85.1 \\
    + DAPT~\cite{zhou2024dynamic} & CVPR 24 & 5.65  &83.87 & 85.7\\
    + PointGST (\textbf{ours})& - & \textbf{5.59}  & \textbf{83.98} & \textbf{85.8} \\
    \midrule
    \revise{\textcolor{gray}{PointGPT-L~\cite{chen2024pointgpt}~(baseline)}} & \revise{\textcolor{gray}{NeurIPS 23}} & \revise{\textcolor{gray}{339.39}} & \revise{\textcolor{gray}{84.80}} & \revise{\textcolor{gray}{86.6}} \\
    \revise{+ IDPT~\cite{zha2023instance}} & \revise{ICCV 23} & \revise{33.17} & \revise{82.91} & \revise{85.1} \\
    \revise{+ Point-PEFT~\cite{tang2024point}} & \revise{AAAI 24} & \revise{31.94}  & \revise{82.34}  &  \revise{84.9} \\
    \revise{+ DAPT~\cite{zhou2024dynamic}} & \revise{CVPR 24} & \revise{33.85} & \revise{83.03} & \revise{85.2} \\
    \revise{+ PointGST (\textbf{ours})} & \revise{-} & \revise{\textbf{31.85}} & \revise{\textbf{83.91}} & \revise{\textbf{85.8}} \\
    \bottomrule
    \end{tabular}

\end{table}

\subsection{Few-shot Learning}
Few-shot learning plays a crucial role in evaluating the efficiency of data usage for training. We make few-shot comparisons on ModelNet40 and the results are listed in Tab.~\ref{tab:fewshot}. Our PointGST beats the previous PEFT methods~\cite{zha2023instance,tang2024point,zhou2024dynamic} in all conducted settings. To further verify the few-shot training ability of our method, we conducted experiments on the challenging ScanObjectNN PB\_T50\_RS dataset, where we set the range from 2\% to 100\%, as depicted in Fig.~\ref{fig:few_show}. Specifically, our PointGST demonstrates strong robustness when trained on limited data. Especially with only 2\% of the training data, PointGST achieves an overall accuracy of 81.96\%, significantly outperforming IDPT~\cite{zha2023instance}, Point-PEFT~\cite{tang2024point}, and DAPT~\cite{zhou2024dynamic} by 1.74\%, 4.34\%, and 4.51\%, respectively. In addition, we report the least trainable parameters, only 0.6M. The results indicate that PointGST efficiently leverages limited training data to capture the underlying characteristics of point clouds, outperforming existing point cloud PEFT methods, particularly in data-constrained environments.

\begin{table}[t]
  \centering
  \scriptsize
  \setlength{\tabcolsep}{1.2mm}
  \caption{Semantic segmentation on the S3DIS~\cite{armeni20163d}. The mean accuracy (mAcc) and mean IoU (mIoU) are reported. Params. represents the trainable parameters.}
    \label{tab:semantic_segmentation}
    \begin{tabular}{lcccc}
    \toprule
    Methods & Reference & Params. (M)& mAcc (\%) & mIoU (\%) \\
    \midrule
    \revise{\textcolor{gray}{Point-BERT~\cite{yu2022point} (baseline)}} & \revise{\textcolor{gray}{CVPR 22}} & \revise{\textcolor{gray}{27.02}} & \revise{\textcolor{gray}{70.3}} & \revise{\textcolor{gray}{60.8}} \\ 
    \revise{+ Linear probing} & \revise{-} & \revise{5.20} & \revise{64.1} & \revise{53.9} \\
    \revise{+ IDPT~\cite{zha2023instance}} & \revise{ICCV 23} & \revise{5.64} & \revise{65.2} & \revise{54.2} \\
    \revise{+ Point-PEFT~\cite{tang2024point}} & \revise{AAAI 24} & \revise{5.58} & \revise{66.9} & \revise{56.0} \\
    \revise{+ DAPT~\cite{zhou2024dynamic}} & \revise{CVPR 24} & \revise{5.61} & \revise{67.0} & \revise{56.3} \\
    \revise{+ PointGST (\textbf{ours})} & \revise{-} & \revise{\textbf{5.55}} & \revise{\textbf{68.6}} & \revise{\textbf{58.4}}\\
    \midrule
    \textcolor{gray}{Point-MAE~\cite{pang2022masked} (baseline)} & \textcolor{gray}{ECCV 22} & \textcolor{gray}{27.02} & \textcolor{gray}{69.9} & \textcolor{gray}{60.8} \\ 
    + Linear probing &- &5.20 &63.4 &52.5 \\
    + IDPT~\cite{zha2023instance} & ICCV 23 & 5.64  & 65.0  & 53.1  \\
    + Point-PEFT~\cite{tang2024point} & AAAI 24 & 5.58  &  66.5 &  56.0 \\
    + DAPT~\cite{zhou2024dynamic} & CVPR 24 & 5.61  & 67.2 & 56.2 \\
    + PointGST (\textbf{ours})& - & \textbf{5.55}  & \textbf{68.4} & \textbf{58.6} \\
    \midrule
    \textcolor{gray}{ACT~\cite{dong2023act} (baseline)} & \textcolor{gray}{ICLR 23} &  \textcolor{gray}{27.02} & \textcolor{gray}{71.1} & \textcolor{gray}{61.2} \\
    + Linear probing &- &5.20 &64.1 &52.0 \\
    + IDPT~\cite{zha2023instance} &  ICCV 23 & 5.64 & 64.1  & 52.1 \\
    + Point-PEFT~\cite{tang2024point} & AAAI 24 & 5.58  & 66.0 & 54.6  \\
    + DAPT~\cite{zhou2024dynamic} & CVPR 24 & 5.61  &64.7 & 54.5\\
    + PointGST (\textbf{ours})& - & \textbf{5.55}  & \textbf{67.6} & \textbf{57.4} \\
    \midrule
    \textcolor{gray}{\recon~\cite{qi2023contrast} (baseline)} & \textcolor{gray}{ICML 23} &  \textcolor{gray}{27.02} & \textcolor{gray}{69.7} & \textcolor{gray}{60.8} \\
    + Linear probing &- &5.20 & 64.3 & 51.2 \\
    + IDPT~\cite{zha2023instance} &  ICCV 23 & 5.64 & 62.9  & 50.5 \\
    + Point-PEFT~\cite{tang2024point} & AAAI 24 & 5.58  & 65.8  & 55.8 \\
    + DAPT~\cite{zhou2024dynamic} & CVPR 24 & 5.61  & 66.3 & 56.3 \\
    + PointGST (\textbf{ours})& - & \textbf{5.55}  & \textbf{67.8} & \textbf{57.9}\\
    \midrule
    \revise{\textcolor{gray}{PointGPT-L~\cite{chen2024pointgpt} (baseline)}} & \revise{\textcolor{gray}{NeurIPS 23}} & \revise{\textcolor{gray}{339.30}} & \revise{\textcolor{gray}{70.6}} & \revise{\textcolor{gray}{62.2}} \\
    \revise{+ Linear probing} & \revise{-} & \revise{29.93} & \revise{62.8} & \revise{53.0} \\
    \revise{+ IDPT~\cite{zha2023instance}} & \revise{ICCV 23} & \revise{33.09} & \revise{69.1} & \revise{57.6} \\
    \revise{+ Point-PEFT~\cite{tang2024point}} & \revise{AAAI 24 }& \revise{31.86}  & \revise{68.4}  & \revise{58.0} \\
    \revise{+ DAPT~\cite{zhou2024dynamic}} & \revise{CVPR 24} & \revise{33.76} & \revise{68.7} & \revise{59.0} \\
    \revise{+ PointGST (\textbf{ours})} & \revise{-} & \revise{\textbf{31.76}} & \revise{\textbf{69.1}} & \revise{\textbf{59.6}}\\
    \bottomrule
    \end{tabular}
\vspace{-10pt}
\end{table}

\begin{table}[h]
  \centering
  \scriptsize
  \setlength{\tabcolsep}{1.mm}
  \caption{  \revise{Scene-level object detection on the ScanNetV2~\cite{dai2017scannet}. The average precision with 3D IoU thresholds of 0.25 ($AP_{25}$) and 0.5 ($AP_{50}$) are reported. Params. represents the trainable parameters.}}
    \label{tab: detection}
    \begin{tabular}{lcccc}
    \toprule
    \revise{Methods} & \revise{Reference} & \revise{Params. (M)}& \revise{$AP_{25}$ (\%)} & \revise{$AP_{50}$ (\%)} \\
    \midrule
    \revise{\textcolor{gray}{Point-BERT~\cite{yu2022point} (baseline)}}& \revise{\textcolor{gray}{CVPR 22}} & \revise{\textcolor{gray}{34.63}} & \revise{\textcolor{gray}{61.0}} & \revise{\textcolor{gray}{38.3}} \\ 
    \revise{+ Linear probing} & \revise{-} & \revise{13.35} & \revise{56.9} & \revise{35.5}  \\
    \revise{+ IDPT~\cite{zha2023instance}} & \revise{ICCV 23} & \revise{14.68}  &  \revise{58.3} & \revise{36.5}  \\
    \revise{+ Point-PEFT~\cite{tang2024point}} & \revise{AAAI 24} & \revise{13.81} & \revise{58.1} & \revise{36.2} \\
    \revise{+ DAPT~\cite{zhou2024dynamic}} & \revise{CVPR 24} & \revise{13.99} & \revise{58.3} & \revise{37.0} \\
    \revise{+ PointGST (\textbf{ours})}& \revise{-} & \revise{\textbf{13.71}}  & \revise{\textbf{58.4}} & \revise{\textbf{38.6}} \\
    \midrule
    \revise{\textcolor{gray}{Point-MAE~\cite{pang2022masked} (baseline)}}& \revise{\textcolor{gray}{ECCV 22}} & \revise{\textcolor{gray}{34.63}} & \revise{\textcolor{gray}{59.3}} & \revise{\textcolor{gray}{39.8}} \\ 
    \revise{+ Linear probing} & \revise{-} & \revise{13.35} & \revise{58.9} & \revise{38.1}  \\
    \revise{+ IDPT~\cite{zha2023instance}} & \revise{ICCV 23} & \revise{14.68}  & \revise{58.9} & \revise{38.2}  \\
    \revise{+ Point-PEFT~\cite{tang2024point}} & \revise{AAAI 24} & \revise{13.81} & \revise{58.0} & \revise{38.9} \\
    \revise{+ DAPT~\cite{zhou2024dynamic}} & \revise{CVPR 24} & \revise{13.99} & \revise{59.6} & \revise{38.5} \\
    \revise{+ PointGST (\textbf{ours})}& \revise{-} & \revise{\textbf{13.71}}  & \revise{\textbf{59.9}} & \revise{\textbf{39.2}} \\
    \midrule
    \revise{\textcolor{gray}{ACT~\cite{dong2023act} (baseline)}}& \revise{\textcolor{gray}{ICLR 23}} & \revise{\textcolor{gray}{34.63}} & \revise{\textcolor{gray}{59.9}} & \revise{\textcolor{gray}{39.8}} \\ 
    \revise{+ Linear probing} & \revise{-} & \revise{13.35} & \revise{56.2} & \revise{36.2} \\
    \revise{+ IDPT~\cite{zha2023instance}} & \revise{ICCV 23} & \revise{14.68}  &  \revise{58.1} & \revise{36.2}  \\
    \revise{+ Point-PEFT~\cite{tang2024point}} & \revise{AAAI 24} & \revise{13.81} & \revise{57.5} & \revise{36.0} \\
    \revise{+ DAPT~\cite{zhou2024dynamic}} & \revise{CVPR 24} & \revise{13.99} & \revise{57.6} & \revise{37.9} \\
    \revise{+ PointGST (\textbf{ours})}& \revise{-} & \revise{\textbf{13.71}}  & \revise{\textbf{60.3}} & \revise{\textbf{40.3}} \\
    \midrule
    \revise{\textcolor{gray}{\recon~\cite{qi2023contrast} (baseline)}}& \revise{\textcolor{gray}{ICML 23}} & \revise{\textcolor{gray}{34.63}} & \revise{\textcolor{gray}{59.6}} & \revise{\textcolor{gray}{40.0}} \\ 
    \revise{+ Linear probing} & \revise{-} & \revise{13.35} & \revise{57.6} & \revise{37.6}  \\
    \revise{+ IDPT~\cite{zha2023instance}} & \revise{ICCV 23} & \revise{14.68}  & \revise{57.4} & \revise{36.2}  \\
    \revise{+ Point-PEFT~\cite{tang2024point}} & \revise{AAAI 24} & \revise{13.81} & \revise{58.5} & \revise{38.2} \\
    \revise{+ DAPT~\cite{zhou2024dynamic}} & \revise{CVPR 24} & \revise{13.99} & \revise{\textbf{59.7}} & \revise{38.7} \\
    \revise{+ PointGST (\textbf{ours})}& \revise{-} & \revise{\textbf{13.71}}  & \revise{59.5} & \revise{\textbf{39.7}} \\
    \midrule
    \revise{\textcolor{gray}{PointGPT-L~\cite{chen2024pointgpt} (baseline)}}& \revise{\textcolor{gray}{NeurIPS 23}} & \revise{\textcolor{gray}{317.72}} & \revise{\textcolor{gray}{60.9}} & \revise{\textcolor{gray}{42.2}} \\ 
    \revise{+ Linear probing} & \revise{-} & \revise{15.48} & \revise{59.9} & \revise{39.1}  \\
    \revise{+ IDPT~\cite{zha2023instance}} & \revise{ICCV 23} & \revise{24.93}  & \revise{60.0} & \revise{41.7}  \\
    \revise{+ Point-PEFT~\cite{tang2024point}} & \revise{AAAI 24} & \revise{17.71} & \revise{60.2} & \revise{41.1} \\
    \revise{+ DAPT~\cite{zhou2024dynamic}} & \revise{CVPR 24} & \revise{18.88} & \revise{60.7} & \revise{40.3} \\
    \revise{+ PointGST (\textbf{ours})}& \revise{-} & \revise{\textbf{17.31}}  & \revise{\textbf{61.2}} & \revise{\textbf{42.2}} \\
    \bottomrule
    \end{tabular}

\end{table}

\subsection{Part Segmentation}
Part segmentation presents challenges in accurately predicting detailed class labels for each point. We apply the Point-BERT~\cite{yu2022point}, Point-MAE~\cite{pang2022masked}, \revise{ACT~\cite{dong2023act}, \recon~\cite{qi2023contrast}, and PointGPT-L~\cite{chen2024pointgpt}} as baselines on the ShapeNetPart dataset. We incorporate the proposed point cloud spectral adapter (PCSA) into every layer of the models. The quantitative results are shown in Tab.~\ref{tab:segmentation}. Our method achieves highly competitive performance while utilizing significantly fewer trainable parameters than fully fine-tuning counterparts (i.e., baseline methods). Notably, the increase in trainable parameters is mainly due to the head part, but we still reduce overall trainable parameters and surpass other PEFT methods like IDPT~\cite{zha2023instance}, Point-PEFT~\cite{tang2024point} and DAPT~\cite{zhou2024dynamic}.

\begin{figure*}[t]
	\begin{center}
		\includegraphics[width=1\linewidth]{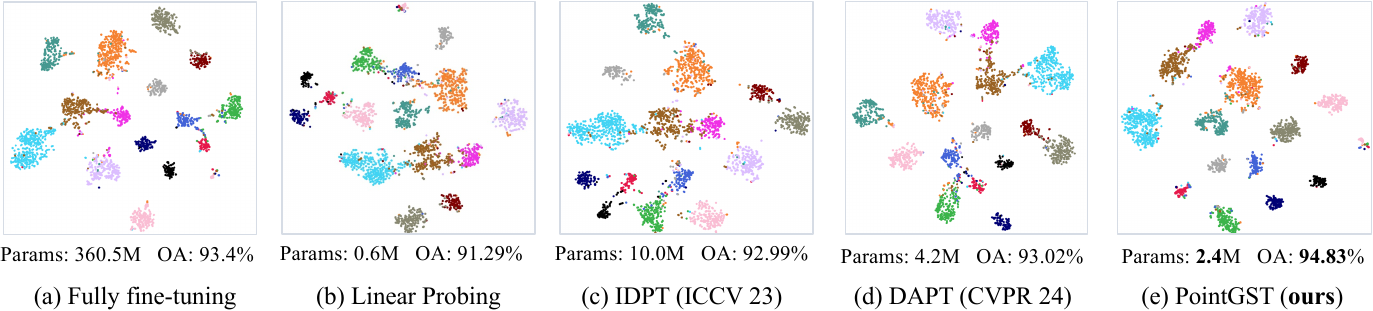}
	\end{center}
    \vspace{-12pt}
	\caption{t-SNE visualizations, along with trainable parameters (Params.) and overall accuracy (OA), are based on the ScanObjectNN PB\_T50\_RS dataset using a pre-trained PointGPT-L with different tuning strategies.}
	\label{fig:tsne}
\end{figure*}

\subsection{Scene-level Semantic Segmentation}

We evaluate the proposed PointGST on the semantic segmentation task using the S3DIS dataset~\cite{armeni20163d}, as shown in Tab.~\ref{tab:semantic_segmentation}. Our empirical observations suggest that previous PEFT methods~\cite{zha2023instance,tang2024point,zhou2024dynamic} struggle with this task, likely due to the models being pre-trained on object-level datasets~\cite{chang2015shapenet}. This results in sub-optimal performance when fine-tuned on a scene-level dataset. In contrast, our PointGST effectively utilizes intrinsic information during fine-tuning, bridging the performance gap between fully and efficiently fine-tuning strategies. For example, using ACT~\cite{dong2023act} as the baseline, PointGST achieves 67.6\% mAcc and 57.4\% mIoU with only 5.55M trainable parameters, substantially outperforming IDPT~\cite{zha2023instance}, Point-PEFT~\cite{tang2024point} and DAPT~\cite{zhou2024dynamic} by 3.5\%/5.3\%, 1.6\%/2.8\%, and 2.9\%/2.9\% in mAcc/mIoU, respectively. Similarly, when applied to \revise{Point-BERT~\cite{yu2022point},} Point-MAE~\cite{pang2022masked}, \recon~\cite{qi2023contrast}, \revise{and PointGPT-L~\cite{chen2024pointgpt},} PointGST consistently surpasses IDPT, Point-PEFT, and DAPT. 

\revise{
\subsection{Scene-level Object Detection}
Following the representative self-supervised methods~\cite{xie2020pointcontrast,yu2022point,pang2022masked,dong2023act}, we conduct object detection on the ScanNetV2 dataset~\cite{dai2017scannet}. Specifically, we replace the default encoder of a standard detector 3DETR~\cite{misra2021end} with the pre-trained backbone, and the results are presented in Tab.~\ref{tab: detection}. The proposed PointGST outperforms all previous point cloud PEFT methods~\cite{zha2023instance,tang2024point,zhou2024dynamic} under different baselines while with less trainable parameters. For instance, when applying ACT~\cite{dong2023act} as the baseline, our PointGST significantly surpasses the IDPT~\cite{zha2023instance}, Point-PEFT~\cite{tang2024point}, and DAPT~\cite{zhou2024dynamic} by 4.1\%, 4.3\%, and 2.4\% under the strict metric ($AP_{50}$) while using less trainable parameters. A similar phenomenon is also presented in the other baselines. These impressive results confirm the generalization ability of the proposed method on the scene detection task. 
}

\Revise{
\subsection{Point Cloud Completion}

To evaluate the generalizability and parameter efficiency of our method in generative tasks, we conduct experiments on point cloud completion using the PCN dataset~\cite{yuan2018pcn}. We adopt AdaPoinTr~\cite{yu2023adapointr} as the pre-trained backbone (using the weights trained on the Projected-ShapeNet-55~\cite{yu2023adapointr} dataset) and compare PointGST with representative parameter-efficient tuning methods~\cite{zha2023instance,zhou2024dynamic,tang2024point}. As shown in Tab.~\ref{tab:completion}, the proposed PointGST achieves the best average Chamfer Distance (CD-$\ell_1$) and matches the highest F-Score@1\% with less trainable parameters. These results demonstrate that our spectral fine-tuning framework excels in discriminative tasks and is also effective for generative tasks such as point cloud completion.
}

\begin{table}[!t]
\scriptsize
\setlength{\tabcolsep}{0.8mm}
\centering
\caption{\Revise{Point cloud completion on the PCN~\cite{yuan2018pcn} dataset. $\text{Avg CD-}\ell_1$ and F-Score@$1\%$ are reported. Params. indicates the trainable parameters from the backbone.}}
\label{tab:completion}
\begin{tabular}{ lcccc }
\toprule
    Methods & Reference & Params. (M) & $\text{Avg CD-}\ell_1$$\downarrow$& F-Score@$1\%$$\uparrow$\\
 \midrule
    \textcolor{gray}{AdaPoinTr~\cite{yu2023adapointr} (baseline)}& \textcolor{gray}{TPAMI 23} & \textcolor{gray}{17.1} & \textcolor{gray}{6.45} & \textcolor{gray}{0.844}  \\
    + Linear probing & - & 0.4 & 6.87 & 0.822  \\
    + IDPT~\cite{zha2023instance} & ICCV 23 & 1.7 & 6.75 & 0.831  \\
    + Point-PEFT~\cite{tang2024point} & AAAI 24 & \textbf{0.6} & 6.74 & 0.828  \\
    + DAPT~\cite{zhou2024dynamic} & CVPR 24 & 0.9 & 6.65 & \textbf{0.836} \\
    + PointGST (\textbf{ours}) & - & \textbf{0.6} & \textbf{6.64} & \textbf{0.836}  \\
\bottomrule
\end{tabular}
\end{table}

\subsection{Compared with PEFT Methods from Other Domains}

Tab.~\ref{tab:other_peft_compare} presents a comparative analysis of various PEFT methods originating from natural language processing (NLP) and 2D vision. The results clearly show that although PEFT methods in other areas can be adapted for 3D tasks, their performance falls short compared to methods specifically designed for 3D vision. For instance, the best-performing PEFT methods from NLP and 2D vision achieve overall accuracies of 83.93\% and 83.66\%, respectively, leaving a significant performance gap compared to fully fine-tuning, which achieves 85.18\%. This indicates the need for 3D-specific adaptations or entirely new methods tailored to 3D data. The existing point cloud PEFT methods, IDPT~\cite{zha2023instance} and DAPT~\cite{zhou2024dynamic} show better performance with accuracies of 84.94\% and 85.08\%, respectively. However, their trainable parameters are noticeably larger than NLP and 2D counterparts and cannot generalize well in different point cloud datasets or pre-trained models (see Sec.~\ref{sec:comparison_dapt_idpt}). In contrast, the proposed PointGST outperforms all other methods with an accuracy of 85.29\%, surpassing IDPT, Point-PEFT~\cite{tang2024point}, and DAPT, with only 0.6M trainable parameters. 

\begin{table}[!t]
\scriptsize
\setlength{\tabcolsep}{0.9mm}
\centering
\caption{Comparisons of Parameter-Efficient Fine-Tuning (PEFT) methods from NLP and 2D vision on the hardest variant of ScanObjectNN~\cite{uy2019revisiting}. Overall accuracy (\%) without voting is reported. Params. is the trainable parameters.}
\label{tab:other_peft_compare}
\begin{tabular}{ lcccc }
\toprule
 Method &Reference & Design for & Params. (M) & PB\_T50\_RS \\
\midrule
 \textcolor{gray}{Point-MAE~\cite{pang2022masked} (baseline)}  &\textcolor{gray}{ECCV 22}& \textcolor{gray}{-} & \textcolor{gray}{22.1} & \textcolor{gray}{85.18}  \\
Linear probing &- &-& 0.3& 75.99\\
 \midrule
  + Adapter~\cite{houlsby2019parameter}&ICML 19 &NLP & 0.9 & 83.93 \\
  + Prefix tuning~\cite{li2021prefix}& ACL 21 &NLP &0.7 & 77.72  \\
  + BitFit~\cite{zaken2022bitfit} & ACL 21 &NLP &0.3 & 82.62    \\
  + LoRA~\cite{hu2021lora} & ICLR 22 &NLP & 0.9&  81.74   \\
  + DEPT~\cite{shi2024dept} & ICLR 24&NLP  &0.3 & 79.70\\
  + FourierFT~\cite{gao2024parameter} & ICML24&NLP  &0.3 & 78.57\\
  \midrule
  + VPT-Deep~\cite{jia2022visual}&ECCV 22&2D  &0.4 &  81.09 \\
  + AdaptFormer~\cite{chen2022adaptformer} &NeurIPS 22&2D  &0.9  & 83.45 \\
  + SSF~\cite{lian2022scaling} & NeurIPS 22&2D   &0.4  & 82.58\\
  + FacT~\cite{jie2023fact} & AAAI 23&2D  & 0.5 &78.76\\
  + BI-AdaptFormer~\cite{jie2023revisiting} & ICCV 23&2D   &0.4 & 83.66\\
  + SCT~\cite{zhao2023sct} & IJCV 24&2D   &0.3 & 80.40\\
 
  \midrule
  + IDPT~\cite{zha2023instance} &ICCV 23&3D  & 1.7 &84.94\\
  + DAPT~\cite{zhou2024dynamic} &CVPR 24&3D  & 1.1 &85.08\\
  + PointGST (\textbf{ours}) & - &3D& 0.6 & \textbf{85.29} \\
\bottomrule
\end{tabular}
\end{table}

\begin{figure}[t]
	\begin{center}
		\includegraphics[width=0.98\linewidth]{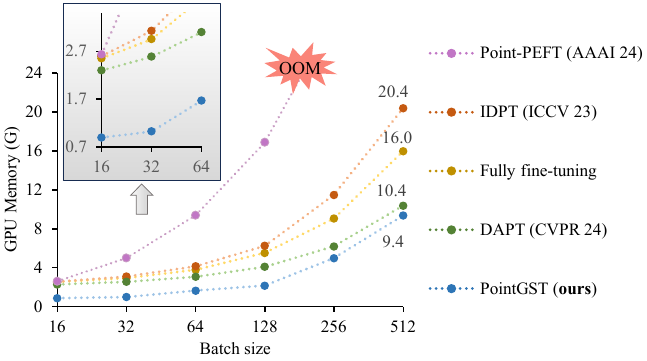}
	\end{center}
    \vspace{-12pt}
	\caption{The comparison of used training GPU memory on NVIDIA 4090.}
	\label{fig:gpu_memory}
\end{figure}

\begin{table*}[!t]
\scriptsize
\setlength{\tabcolsep}{3.5mm}
\centering
\caption{The comparison of our PointGST and baseline methods in terms of storage usage when fine-tuning various downstream point cloud datasets or tasks. }
\label{tab:storage}

\begin{tabular}{ cccccccccc }
   \toprule
 Method & OBJ\_BG & OBJ\_ONLY &PB\_T50\_RS & ModelNet40 & ModelNet Few-shot & ShapeNetPart &S3DIS & Total \\
\midrule
Point-MAE~\cite{pang2022masked}& 22.1M & 22.1M & 22.1M & 22.1M & 22.1M & 27.1M & 27.0M &164.6M\\
PointGST (ours) & 0.6M & 0.6M & 0.6M & 0.6M& 0.6M & 5.6M & 5.6M &\textbf{14.2M} \\
\midrule
PointGPT-L~\cite{chen2024pointgpt} & 360.5M &360.5M & 360.5M & 360.5M & 360.5M & 339.4M & 339.3M & 2481.2M\\
PointGST (ours) & 2.4M & 2.4M & 2.4M & 2.4M & 2.4M &31.9M &31.8M &\textbf{75.7M} \\
\bottomrule
\end{tabular}
\end{table*}

\begin{table}[t]
  \centering
  \scriptsize
  \setlength{\tabcolsep}{0.7mm}
  \caption{\revise{Comparison of different fine-tuning strategies on GPU consuming and inference speed (FPS). We report the results with a batch size of 8 on a single NVIDIA RTX 4090. }}
    \label{tab:costs_comparision}
    \begin{tabular}{lcccccc}
    \toprule
    \revise{Methods} & \revise{Reference} & \revise{Params. (M)} & \revise{Infer. memory (G)}  & \revise{FPS} & \revise{PB\_T50\_RS} \\
    \midrule
    \revise{\textcolor{gray}{PointGPT-L~\cite{chen2024pointgpt}}} & \revise{\textcolor{gray}{NeurIPS 23}} & \revise{\textcolor{gray}{360.5}}  & \revise{\textcolor{gray}{3.8}} & \revise{\textcolor{gray}{72.9}} & \revise{\textcolor{gray}{93.40}} \\
    \revise{IDPT~\cite{zha2023instance}} & \revise{ICCV 23} & \revise{10.0} & \revise{3.9} & \revise{70.4} & \revise{92.99\dtplus{-0.41}} \\
    \revise{Point-PEFT~\cite{tang2024point}} & \revise{AAAI 24} & \revise{3.1} & \revise{4.0} & \revise{32.4} & \revise{92.85\dtplus{-0.55}} \\
    \revise{DAPT~\cite{zhou2024dynamic}} & \revise{CVPR 24} & \revise{4.2} & \revise{3.9} & \revise{66.7} & \revise{93.02\dtplus{-0.38}} \\
    \revise{PointGST (\textbf{ours})} & \revise{-} & \revise{\textbf{2.4}} & \revise{\textbf{3.9}} & \revise{\textbf{71.3}} & \revise{\textbf{94.83}\dplus{+1.43}} \\
    \bottomrule
    \end{tabular}

\end{table}

\begin{table}[!t]
\centering
\scriptsize
\setlength{\tabcolsep}{1.0mm}
\caption{\Revise{Inference time and GPU memory percentages for different components of our PointGST on ScanNetV2~\cite{dai2017scannet} dataset (NVIDIA RTX 4090, batch size 1). }}
\label{tab:total_time}
\begin{tabular}{lccccl}
\toprule
\multirow{2.3}{*}{Metric}  & \multicolumn{2}{c}{Pre-processing} & \multirow{2.3}{*}{Others} & \multirow{2.3}{*}{Total} \\
\cmidrule(lr){2-3}
 & Graph construction & Eigen-decomposition &  &  \\
\midrule
Inference time (\%) & 0.42 & 1.60 & 97.98 & 100 \\
GPU memory (\%)     & 5.60 & 0.58 & 93.82 & 100 \\
\bottomrule
\end{tabular}
\end{table}

\subsection{Visualization Results}
Fig.~\ref{fig:tsne} presents the t-SNE~\cite{van2008visualizing} feature manifold visualization of models after full fine-tuning, linear probing, and point cloud-customized PEFT methods, IDPT~\cite{zha2023instance} and DAPT~\cite{zhou2024dynamic}, alongside our PointGST, applied to the ScanObjectNN PB\_T50\_RS dataset. Greater dispersion of points among categories indicates better model representation and easier classification. Specifically, as shown in Fig.~\ref{fig:tsne}(c-d), the spatial domain tuning-based methods, IDPT and  DAPT exhibit greater confusion between categories, while Fig.~\ref{fig:tsne}(e) demonstrates that our PointGST enables the pre-trained model to produce more distinguishable representations with fewer learnable parameters in the spectral domain.

\subsection{Training and \revise{Inference} Cost}
Our method requires only fine-tuning the proposed lightweight Point Cloud Spectral Adapter (PCSA) while freezing the entire pre-trained backbone. For example, we just fine-tune about 0.67\% of the parameters in the SOTA baseline, PointGPT-L~\cite{chen2024pointgpt}, achieving promising results. As the number of downstream tasks or datasets requiring fine-tuning increases, we only need to store the task/dataset-specific PCSA and one original pre-trained model. As shown in Tab.~\ref{tab:storage}, when fine-tuning various downstream tasks, our PointGST significantly compresses model size compared to the baselines~\cite{pang2022masked,chen2024pointgpt}. 

We further analyze the training GPU memory, an important metric for evaluating the training cost. As shown in Fig.~\ref{fig:gpu_memory}, \revise{the proposed PointGST is highly efficient, consuming considerably less GPU memory than the fully fine-tuned models. Even compared to other PEFT methods such as IDPT~\cite{zha2023instance}, DAPT~\cite{zhou2024dynamic}, and Point-PEFT~\cite{tang2024point}, our method performs better in terms of memory efficiency. Notably, methods like Point-PEFT and IDPT cause significant memory consumption spikes. For example, Point-PEFT experiences a substantial memory increase when scaling batch size, which may be problematic for GPUs with limited memory (e.g., 24 GB cards). During the testing phase, as shown in Tab.~\ref{tab:costs_comparision}, while the inference speed (FPS) of PEFT methods, including ours, is slightly lower than that of the baseline (FFT), our method demonstrates the best FPS among all PEFT approaches. This is expected, as adapter-tuning-based methods typically freeze the pre-trained backbone and introduce lightweight learnable modules, which slightly impact inference speed due to additional operations. Besides, the inference GPU memory of our method is highly competitive with the FFT and other PEFT methods.} 

\Revise{We also provide a detailed analysis of both inference time and GPU memory usage percentages for different components of PointGST on the large-scale ScanNetV2 dataset, with PointGPT-L~\cite{chen2024pointgpt} as the baseline, as shown in Tab.~\ref{tab:total_time}. We observe that the percentages of time and GPU memory consumed by graph construction and Laplacian eigen-decomposition constitute only a small fraction of the total inference process, with the vast majority allocated to other standard operations. These results indicate that the additional spectral-domain computations introduced by our method incur negligible overhead in both inference time and GPU memory usage.
}

\begin{table}[!t]
\scriptsize
\setlength{\tabcolsep}{1mm}
\centering
\caption{The effect of each component of our PointGST. The trainable parameters (Params.) and the overall accuracy (\%) are reported. The results before and after ``/" denote using Point-MAE and PointGPT-L as baselines, respectively.}
\label{tab:compoments}
\begin{tabular}{ cccccc }
   \toprule
 Global spectral& Local spectral & Residual & Inverse &  Params. (M) & PB\_T50\_RS \\
\midrule
\multicolumn{4}{c}{Full fine-tuning}  &22.1/360.5 & 85.18/93.40 \\
\multicolumn{4}{c}{Linear Probing}  &0.27/0.60 & 75.99/91.29 \\
\midrule
\checkmark &- & -& -&0.62/2.43 & 81.99/93.88\\
\checkmark &- & \checkmark & - & 0.62/2.43 & 83.03/94.27 \\
\checkmark &- & - & \checkmark & 0.62/2.43 & 84.07/94.48\\
\checkmark &- & \checkmark & \checkmark & 0.62/2.43& 84.52/94.69\\
\midrule
- &\checkmark & -& -& 0.62/2.43 & 81.99/94.17 \\
- &\checkmark & \checkmark & - &0.62/2.43 & 83.38/94.45 \\
- & \checkmark &- & \checkmark & 0.62/2.43 & 83.69/94.55 \\
- &\checkmark & \checkmark & \checkmark & 0.62/2.43 & 83.83/94.62\\
\midrule
 \checkmark & \checkmark &  \checkmark & \checkmark& 0.62/2.43 & \textbf{85.29}/\textbf{94.83}\\
\bottomrule
\end{tabular}
\end{table}

\begin{table}[t]
    \centering
    \scriptsize
    \setlength{\tabcolsep}{2.4mm}
    \caption{\revise{The comparison of different tuning types. The results before and after ``/" denote using Point-MAE and PointGPT-L as baselines, respectively.}}
    \label{tab:tuning_type}
    \begin{tabular}{cccc}
    \toprule
    \revise{Tuning type} & \revise{Method} & \revise{Params. (M)} & \revise{PB\_T50\_RS} \\
    \midrule
    \multirow{4}{*}{\revise{Spatial tuning}} & \revise{Fully fine-tuning} & \revise{22.1/360.5} & \revise{85.18/93.40} \\
    & \revise{IDPT~\cite{zha2023instance}} & \revise{1.7/10.0} & \revise{84.94/92.99} \\
    & \revise{DAPT~\cite{zhou2024dynamic}}  & \revise{1.1/4.2} & \revise{85.08/93.02} \\
    & \revise{PointGST w/o spectral}   & \revise{0.6/2.4} & \revise{84.04/93.56} \\
    \midrule
    \revise{Spectral tuning} & \revise{PointGST (ours)}  & \revise{\textbf{0.6}/\textbf{2.4}} & \revise{\textbf{85.29}/\textbf{94.83}} \\
    \bottomrule 
    \end{tabular}
\end{table}

\begin{table*}[!h]
    \centering
    \scriptsize
    \caption{Ablation study on the Point Cloud Spectral Adapter (PCSA), including dimension $r$, scale $s$, and the effect of different adaption choices. The trainable parameters (Params.) and the overall accuracy (\%) are reported. The results before and after ``/" denote using Point-MAE and PointGPT-L as baselines, respectively.}
    \label{tab: design_adapter}
    \resizebox{1.\linewidth}{!}{
    \begin{subtable}{0.33\linewidth} 
        \centering
        \scriptsize
        \setlength{\tabcolsep}{3mm} 
        \caption{The effect of $r$.}
        \label{tab:rank}
        \begin{tabular}{ccc}
            \toprule
            Dimension $r$& Params. (M) & PB\_T50\_RS \\
            \midrule
            12 & 0.39/1.22 & 83.66/94.83  \\
            24 & 0.50/1.82 & 84.59/94.41 \\
            36 & 0.62/2.43 & \textbf{85.29}/\textbf{94.83} \\
            48 & 0.75/3.05 & 84.73/94.41 \\
            72 & 1.00/4.30 & 84.28/94.34 \\
            96 & 1.28/5.58 & 84.98/94.59 \\
            \bottomrule
        \end{tabular}
    \end{subtable}

    \begin{subtable}{0.2\linewidth} 
        \centering
        \scriptsize
         \setlength{\tabcolsep}{2mm} 
        \caption{The effect of scale $s$.}
        \label{tab:scale}
        \begin{tabular}{cc}
            \toprule
             Scale  &PB\_T50\_RS \\
            \midrule
              0.01  &  83.41/94.48 \\
              0.1   &  83.34/94.80 \\
              1.0 & \textbf{85.29}/\textbf{94.83} \\
              2.0 & 83.62/94.41 \\
              5.0 & 71.03/93.72 \\
              10.0 & 67.52/93.65 \\
            \bottomrule
        \end{tabular}
    \end{subtable}

    \begin{subtable}{0.38\textwidth}
        \centering
        \renewcommand\arraystretch{1.1666}
        \setlength{\tabcolsep}{3.7mm}
        \caption{The effect of different adaption choices.}
        \label{tab:different_learnable_architectures}
        \begin{tabular}{ccc}
            \toprule
            Module  & Params. (M)&PB\_T50\_RS \\
            \midrule
            Shared Attention   &0.67/2.52 & 84.07/94.76 \\
            Shared MLP & 0.67/2.52 & 83.69/94.41\\
            Shared DWConv &0.61/2.41 & 85.15/94.55 \\
            \hline
            Shared linear &0.62/2.43 & \textbf{85.29}/\textbf{94.83} \\
            Independent linear &0.64/2.46 & 84.80/94.51 \\
            \bottomrule 
        \end{tabular}
    \end{subtable}
}
\end{table*}

\subsection{Ablation Study}
Following previous methods~\cite{zhou2024dynamic,zha2023instance}, we conduct ablation studies on the challenging ScanObjectNN PB\_T50\_RS dataset. Unless otherwise specified, we choose Point-MAE~\cite{pang2022masked} and PointGPT-L~\cite{chen2024pointgpt} as the baseline, and all experiments are performed without the voting strategy.

\subsubsection{The effect of each component}

We systematically evaluate the effect of each component within our PointGST by analyzing four key aspects: global spectral, local spectral, residual connections, and spectral inversion. The results are detailed in Tab.~\ref{tab:compoments}.

Taking the Point-MAE baseline as an example, without any of our components, the linear probing setting achieves only 75.99\% overall accuracy (OA), significantly lower than the fully fine-tuning counterpart by 9.19\%. When incorporating either global or local spectral components individually, we observe a marked improvement of 6.00\% in OA for both, while introducing only 0.35M additional trainable parameters. Further performance gains are seen when integrating residual connections and applying the inverse Graph Fourier Transform (iGFT) to project back into the spatial domain, aligning with the output of the transformer layer. Ultimately, the combination of all components leads to a significant increase in OA to 85.29\%. This outcome underscores the complementary nature of global and local spectral representations, which can effectively de-correlate the inner confusion of point tokens and provide task-related intrinsic information to guide the fine-tuning phase.

\revise{
To make a more direct comparison between spatial tuning and spectral tuning, we provide additional results where our method does not transform tokens into the spectral domain but instead processes the original spatial tokens directly. As shown in Tab.~\ref{tab:tuning_type}, we observe that the proposed method achieves better performance compared with the spatial domain-tuning methods, demonstrating the importance of fine-tuning in the spectral domain. It also highlights the benefits of token de-correlation and intrinsic task-specific information provided by the spectral basis.
}

\subsubsection{Ablation study on the hyper-parameters and design choices of PCSA}

\textbf{The effect of dimension $r$.} This parameter is used to balance efficiency and performance. As shown in Tab.~\ref{tab:rank}, as the dimension $r$ increases from 12 to 36, the performance of our method improves, with performance increasing from 83.66 to 85.29. However, further increasing $r$ to 96 slightly hurts the performance, with performance dropping to 84.98. Therefore, we set the dimension $r$ to 36 by default, which reports ideal performance and less trainable parameters.

\textbf{The effect of scale $s$.} This parameter controls the impact of spectral tokens on the task-agnostic tokens. As shown in Tab.~\ref{tab:scale}, we empirically find that when the value is simply set as $1$ without the need for extensive hyper-parameter tuning, the proposed method can achieve satisfactory performance.

\textbf{The effect of different adaption choices.} In our PCSA, we utilize a simple shared linear layer to adapt the global and local spectral tokens, shown in Fig.~\ref{fig:pipeline}(b). Here, we explore other possible learnable choices, including shared Attention, MLP (two layers), DWConv, linear, and independent linear layers for global and local spectral. Note that for fair comparisons, we make these options meet similar trainable parameters (about 0.6M). The results presented in Tab.~\ref{tab:different_learnable_architectures} validate that the used shared linear layer is simple yet effective compared with other learnable architectures.

\begin{table}[t]
    \centering
    \scriptsize
    \caption{The effect of inserted position. The trainable parameters (Params.) and the overall accuracy (\%) are reported. The results before and after ``/" denote using Point-MAE and PointGPT-L as baselines, respectively.}
    \label{tab:postion}
    \resizebox{1\linewidth}{!}{
    \begin{subtable}{0.66\linewidth} 
    \setlength{\tabcolsep}{0.4mm} 
        \centering
        \small
        \caption{The effect of inserted layers.}
        \label{tab:inserted_layer}
        \begin{tabular}{ccc}
            \toprule
            Layers &  Params. (M)  & PB\_T50\_RS \\
            \midrule
            1$\rightarrow$6/1$\rightarrow$12  & 0.45/1.51 &83.17/94.69\\
            7$\rightarrow$12/13$\rightarrow$24 & 0.45/1.51 &83.76/94.07\\
             1$\rightarrow$12/1$\rightarrow$24  &0.62/2.43 & \textbf{85.29}/\textbf{94.83}\\
            \bottomrule 
        \end{tabular}
    \end{subtable}

    \begin{subtable}{0.29\textwidth}
    \centering
    \small
    \setlength{\tabcolsep}{0.6mm}
    \caption{The effect of inserted module.}
    \label{tab:inserted_module}
    \begin{tabular}{cc}
        \toprule
        Position  & PB\_T50\_RS \\
        \midrule
        Attention   & 84.28/94.41 \\
         FFN & \textbf{85.29}/\textbf{94.83} \\
        Entire transformer layer & 84.25/94.38 \\
        \bottomrule 
    \end{tabular}
    \end{subtable}
    }
\end{table}

\subsubsection{The effect of the position of PCSA}

We then study the effect of the position of the PCSA. For the inserted layers, we consider the shallow layers and deep layers. Tab.~\ref{tab:inserted_layer} clearly demonstrates that the performance of PointGST has a positive relative to the added layers. Taking the Point-MAE baseline as an example, injecting the PCSA into half of the layers will cause about a 2.12\%/1.53\% performance drop. Therefore, in our method, we inject the proposed point spectral adapter into all transformer layers. In addition, we also explore the effect of inserted modules. There are three potential choices: FFN, Attention, and each transformer block. As shown in Tab.~\ref{tab:inserted_module}, inserting the PCSA into the FFN outperforms Attention and the entire transformer block by 1.01\% and 1.04\% overall accuracy for Point-MAE, respectively.

\subsubsection{Analysis on the point cloud graph construction}

\textbf{The effect of different weighting schemes.} To calculate the edge weight between two points in a graph, we employ a data-dependent scaling strategy. Here, we study various weighting schemes, as detailed in Tab.~\ref{tab:graph_form}. The most straightforward weighting scheme, Euclidean distance, yields sub-optimal performance with an overall accuracy of 85.15\% and 94.73\%. As discussed in Sec.~\ref{sec:point2graph}, we contend that the closer two points are, the stronger the weight of their relationship should be. Therefore, we propose a data-dependent scaled distance for edge weight, which outperforms Euclidean distance by 0.14\% and 0.10\% on the Point-MAE and PointGPT-L baselines. Additionally, we evaluate cosine distance and Gaussian similarity~\cite{von2007tutorial} for each pair of key points, finding that our proposed data-dependent scaling strategy yields more stable results.

We also try to generate the weights between point tokens instead of the original point cloud. The results of the last two lines indicate poor performance, accompanying extremely high training costs (about 4 NVIDIA 3090 days). It is reasonable, as the encoded point tokens from the frozen models are hard to bring the intrinsic information of the downstream point cloud. In contrast, constructing the graph from the original downstream point cloud will provide a compact space that involves task-specific information, which effectively achieves targeted tuning.

\textbf{The effect of different sorting methods and group number $k$.} When constructing local sub-graphs, we sort and categorize the sampled key points into $k$ groups. This part evaluates the impact of various sorting methods, including random, k-nearest neighbors (KNN), and space-filling curves (e.g., Hilbert order and Z-order). Tab.~\ref{tab:sort} shows that random sorting provides the least favorable results, while methods like KNN and space-filling curves that preserve spatial locality considerably enhance the performance. Among them, Trans Z-order sorting stands out as the most effective. KNN exhibits inferior performance, potentially because it does not yield stable results and sometimes aggregates identical points into different groups. In contrast, space-filling curves maintain a consistent scanning pattern, making it easier for models to learn from the data. Besides, as shown in Tab.~\ref{tab:group}, we empirically find that the best performance is achieved when $k$ is set as 4.

\setlength\tabcolsep{1.5pt}
\begin{figure*}[ht]
    \centering
    \scriptsize
    \begin{minipage}{0.715\textwidth}
    \captionsetup{type=table}
    \caption{Analysis on the point cloud graph.}
    \label{tab:analysis_on_spectral}
    \resizebox{1.\linewidth}{!}{

    \begin{subtable}{0.406\textwidth}
        \vspace{5pt}
        \renewcommand\arraystretch{1.06}
        \caption{The effect of edge weights.}
        \vspace{5pt}
        \label{tab:graph_form}
        \begin{tabular}{ccc}
            \toprule
        Source & Method  & PB\_T50\_RS \\
            \midrule
        \multirow{4}{*}{Point cloud}&Euclidean Distance & 85.15/94.73 \\
            &Cosine Distance   & 83.97/94.41 \\
            &Gaussian Similarity  & 83.76/94.73 \\
             &Scaling Strategy (ours)& \textbf{85.29}/\textbf{94.83} \\
             \midrule
        \multirow{2}{*}{Point token} &Gaussian Similarity  & 68.67/94.48 \\
        & Scaling Strategy (ours) & 71.17/94.66 \\
            \bottomrule 
        \end{tabular}
    \end{subtable}

    \hspace{0.5pt}

        \begin{subtable}{0.27\textwidth}
        \renewcommand\arraystretch{1.155}
        \caption{The effect of sorting methods.}
        \setlength{\tabcolsep}{1.9mm}
        \label{tab:sort}
        \begin{tabular}{cc}
            \toprule
            Sort & PB\_T50\_RS \\
            \midrule
            Random  &  83.83/94.27 \\
            KNN   &84.59/94.45 \\
            Hilbert  &  84.00/94.45 \\
            Trans Hilbert   & 84.59/94.34 \\
            Z-order  &  84.90/94.24 \\
            Trans Z-order   &\textbf{85.29}/\textbf{94.83} \\
            \bottomrule
        \end{tabular}
    \end{subtable}
    
    \hspace{1.8pt}

    \begin{subtable}{0.25\textwidth}
        \renewcommand\arraystretch{1.35}
            \caption{The effect of group number $k$.}
            \setlength{\tabcolsep}{0.6mm}
            \label{tab:group}
            \begin{tabular}{cccc}
            \toprule
            \textit{k} & PB\_T50\_RS\\ 
            \midrule
            w/o sub-graphs & 84.52/94.69 \\
            $2$ & 84.04/94.45\\
            $4$ & \textbf{85.29}/\textbf{94.83}\\
            $8$ & 84.04/94.52\\
            $16$ & 84.84/94.55\\
            \bottomrule
        \end{tabular}
    \end{subtable}

}    
    \end{minipage}
    \hfill
    \begin{minipage}{0.28\textwidth}
    \captionsetup{type=table}
    \caption{Analysis on the spectral conversion.}
   
            \label{tab:fourier}
            \renewcommand\arraystretch{2.15}
            \resizebox{1.\linewidth}{!}{
            \begin{tabular}{cc}
            \toprule
            Method & PB\_T50\_RS\\
            \midrule
            Without Fourier Transform & 84.04/93.56\\
            Fast Fourier Transform (FFT) & 83.69/94.55 \\
            Discrete Cosine Transform (DCT) & 84.25/94.25\\
            Graph Fourier Transform (GFT) & \textbf{85.29}/\textbf{94.83} \\
            \bottomrule

            \end{tabular}
        }
\end{minipage}
\end{figure*}

\subsubsection{Analysis on Different Spectral Conversion}

In this part, we investigate the effect of different spectral conversions, including the Fast Fourier Transform (FFT), Discrete Cosine Transform (DCT), and we used Graph Fourier Transform (GFT). As shown in Tab.~\ref{tab:fourier}, the results demonstrate the superiority of the GFT we applied. This can be attributed to the fact that the FFT and DCT are designed for regularly sampled or structured data and assume uniform spacing and regularity in the signals, making them unsuitable for processing non-Euclidean point clouds~\cite{hu2021graph}. In contrast, to accommodate the irregular domain of point clouds, we apply the GFT, which is well-suited for such data structures and adapts to specific samples.

\Revise{
\subsubsection{The Effect of the Input Point Number}

Finally, to evaluate the effect of different input point densities, we conduct experiments on ModelNet40~\cite{wu20153d} dataset using 128 (sparse), 1024 (default), and 8192 (dense) points as input, as summarized in Tab.~\ref{tab:different_points}. PointGST consistently achieves the best or most competitive accuracy across all point regimes for both Point-MAE~\cite{pang2022masked} and PointGPT-L~\cite{chen2024pointgpt} baselines. In particular, our method maintains high performance at the default 1024-point setting and exhibits strong robustness under extremely sparse (128) and dense (8192) inputs. For example, with the PointGPT-L baseline, our PointGST attains 91.7\%, 94.8\%, and 94.7\% accuracy at 128, 1024, and 8192 points, respectively, outperforming all other parameter-efficient fine-tuning methods~\cite{zha2023instance,zhou2024dynamic,tang2024point} in each case. This demonstrates that the proposed spectral adapter can effectively adapt to large variations in point cloud density, making our method suitable for diverse real-world scenarios where input may vary.

\begin{table}[!t]
\scriptsize
\setlength{\tabcolsep}{0.5mm}
\centering
\caption{\Revise{Classification accuracy on ModelNet40~\cite{wu20153d} dataset with 128, 1024, and 8192 input points (no voting). Params. denotes the trainable parameters.}}
\label{tab:different_points}
\begin{tabular}{ lcccccc }
\toprule
    Pre-trained model & Fine-tuning strategy & Reference & Params. (M) & 128 & 1024 & 8192 \\
 \midrule
    \multirow{5}{*}{\tabincell{c}{Point-MAE~\cite{pang2022masked}\\(ECCV 22)}} 
    & \textcolor{gray}{Fully fine-tune} & \textcolor{gray}{-} & \textcolor{gray}{22.1} & \textcolor{gray}{91.1} & \textcolor{gray}{93.2} & \textcolor{gray}{93.4} \\
    & IDPT~\cite{zha2023instance} & ICCV 23 & 1.7 & 91.4 & 93.3 & 93.5 \\
    & Point-PEFT~\cite{tang2024point} & AAAI 24 & 0.7 & 91.1 & 93.3 & 93.6 \\
    & DAPT~\cite{zhou2024dynamic} & CVPR 24 & 1.1 & 91.2 & 93.5 & 93.2 \\
    & PointGST (\textbf{ours}) & - & \textbf{0.6} & \textbf{91.5} & \textbf{93.5} & \textbf{93.8} \\
 \midrule
    \multirow{5}{*}{\tabincell{c}{PointGPT-L~\cite{chen2024pointgpt}\\(NeurIPS 23)}} 
    & \textcolor{gray}{Fully fine-tune} & \textcolor{gray}{-} & \textcolor{gray}{360.5} & \textcolor{gray}{91.5} & \textcolor{gray}{94.1} & \textcolor{gray}{92.7} \\
    & IDPT~\cite{zha2023instance} & ICCV 23 & 10.0 & 89.3 & 93.4 & 92.0 \\
    & Point-PEFT~\cite{tang2024point} & AAAI 24 & 3.1 & 90.0 & 93.5 & 94.4 \\
    & DAPT~\cite{zhou2024dynamic} & CVPR 24 & 4.2 & 90.8 & 94.2 & 94.1 \\
    & PointGST (\textbf{ours}) & - & \textbf{2.4} & \textbf{91.7} & \textbf{94.8} & \textbf{94.7} \\
\bottomrule
\end{tabular}
\end{table}

}

\section{Conclusion}

In this paper, we propose a novel parameter-efficient fine-tuning (PEFT) method for point cloud learning called PointGST. Unlike previous spatial domain-tuning point cloud PEFT methods, our PointGST shifts fine-tuning to the spectral domain through the proposed Point Cloud Spectral Adapter (PCSA). This approach effectively mitigates confusion among point tokens from frozen models by using orthogonal components to separate them and incorporates task-specific intrinsic information for targeted tuning. 
Extensive experiments across various challenging datasets demonstrate that PointGST outperforms the fully fine-tuning setting and SOTA point cloud PEFT methods, building new state-of-the-art results. This makes PointGST a practical solution for resource-efficient model adaptation in point cloud tasks. We hope this work can open new directions for efficient fine-tuning in point cloud tasks.

\Revise{
\noindent\textbf{Limitation and Future Work.} Currently, our spectral domain PEFT framework is not directly applicable to more challenging 3D generative tasks (e.g., text-to-3D or image-to-3D generation), which often require fundamentally different architectures and optimization strategies. Extending PointGST to such generative scenarios is a promising direction for future research.}

{\small
\bibliographystyle{IEEEtran}
\bibliography{reference}

\begin{thebibliography}{100}
\providecommand{\url}[1]{#1}
\csname url@samestyle\endcsname
\providecommand{\newblock}{\relax}
\providecommand{\bibinfo}[2]{#2}
\providecommand{\BIBentrySTDinterwordspacing}{\spaceskip=0pt\relax}
\providecommand{\BIBentryALTinterwordstretchfactor}{4}
\providecommand{\BIBentryALTinterwordspacing}{\spaceskip=\fontdimen2\font plus
\BIBentryALTinterwordstretchfactor\fontdimen3\font minus \fontdimen4\font\relax}
\providecommand{\BIBforeignlanguage}[2]{{%
\expandafter\ifx\csname l@#1\endcsname\relax
\typeout{** WARNING: IEEEtran.bst: No hyphenation pattern has been}%
\typeout{** loaded for the language `#1'. Using the pattern for}%
\typeout{** the default language instead.}%
\else
\language=\csname l@#1\endcsname
\fi
#2}}
\providecommand{\BIBdecl}{\relax}
\BIBdecl

\bibitem{li2023pillarnext}
J.~Li, C.~Luo, and X.~Yang, ``Pillarnext: Rethinking network designs for 3d object detection in lidar point clouds,'' in \emph{Proc. IEEE Conf. Comput. Vis. Pattern Recognit.}, 2023, pp. 17\,567--17\,576.

\bibitem{fan2023super}
L.~Fan, Y.~Yang, F.~Wang, N.~Wang, and Z.~Zhang, ``Super sparse 3d object detection,'' \emph{IEEE Trans. Pattern Anal. Mach. Intell.}, vol.~45, no.~10, pp. 12\,490--12\,505, 2023.

\bibitem{melas2023pc2}
L.~Melas-Kyriazi, C.~Rupprecht, and A.~Vedaldi, ``Pc2: Projection-conditioned point cloud diffusion for single-image 3d reconstruction,'' in \emph{Proc. IEEE Conf. Comput. Vis. Pattern Recognit.}, 2023, pp. 12\,923--12\,932.

\bibitem{xiao2023unsupervised}
A.~Xiao, J.~Huang, D.~Guan, X.~Zhang, S.~Lu, and L.~Shao, ``Unsupervised point cloud representation learning with deep neural networks: A survey,'' \emph{IEEE Trans. Pattern Anal. Mach. Intell.}, vol.~45, no.~9, pp. 11\,321--11\,339, 2023.

\bibitem{xu2023pointllm}
R.~Xu, X.~Wang, T.~Wang, Y.~Chen, J.~Pang, and D.~Lin, ``Pointllm: Empowering large language models to understand point clouds,'' in \emph{Proc. Eur. Conf. Comput. Vis.}, 2024, pp. 131--147.

\bibitem{devlin2019bert}
J.~Devlin, M.-W. Chang, K.~Lee, and K.~Toutanova, ``Bert: Pre-training of deep bidirectional transformers for language understanding,'' in \emph{Proc. Annual Meeting of the Association for Computational Linguistics}, 2019, pp. 4171--4186.

\bibitem{he2022masked}
K.~He, X.~Chen, S.~Xie, Y.~Li, P.~Doll{\'a}r, and R.~Girshick, ``Masked autoencoders are scalable vision learners,'' in \emph{Proc. IEEE Conf. Comput. Vis. Pattern Recognit.}, 2022, pp. 16\,000--16\,009.

\bibitem{pang2022masked}
Y.~Pang, W.~Wang, F.~E. Tay, W.~Liu, Y.~Tian, and L.~Yuan, ``Masked autoencoders for point cloud self-supervised learning,'' in \emph{Proc. Eur. Conf. Comput. Vis.}, 2022, pp. 604--621.

\bibitem{yu2022point}
X.~Yu, L.~Tang, Y.~Rao, T.~Huang, J.~Zhou, and J.~Lu, ``Point-bert: Pre-training 3d point cloud transformers with masked point modeling,'' in \emph{Proc. IEEE Conf. Comput. Vis. Pattern Recognit.}, 2022, pp. 19\,313--19\,322.

\bibitem{wang2024point}
Z.~Wang, Y.~Rao, X.~Yu, J.~Zhou, and J.~Lu, ``Point-to-pixel prompting for point cloud analysis with pre-trained image models,'' \emph{IEEE Trans. Pattern Anal. Mach. Intell.}, vol.~46, no.~6, pp. 4381--4397, 2024.

\bibitem{dong2023act}
R.~Dong, Z.~Qi, L.~Zhang, J.~Zhang, J.~Sun, Z.~Ge, L.~Yi, and K.~Ma, ``Autoencoders as cross-modal teachers: Can pretrained 2d image transformers help 3d representation learning?'' in \emph{Proc. Int. Conf. Learn. Representations}, 2023.

\bibitem{zhang2022point}
R.~Zhang, Z.~Guo, P.~Gao, R.~Fang, B.~Zhao, D.~Wang, Y.~Qiao, and H.~Li, ``Point-m2ae: multi-scale masked autoencoders for hierarchical point cloud pre-training,'' in \emph{Proc. Adv. Neural Inf. Process. Syst.}, vol.~35, 2022, pp. 27\,061--27\,074.

\bibitem{chen2024pointgpt}
G.~Chen, M.~Wang, Y.~Yang, K.~Yu, L.~Yuan, and Y.~Yue, ``Pointgpt: Auto-regressively generative pre-training from point clouds,'' in \emph{Proc. Adv. Neural Inf. Process. Syst.}, vol.~36, 2023, pp. 29\,615--29\,627.

\bibitem{liang2024pointmamba}
D.~Liang, X.~Zhou, W.~Xu, X.~Zhu, Z.~Zou, X.~Ye, X.~Tan, and X.~Bai, ``Pointmamba: A simple state space model for point cloud analysis,'' in \emph{Proc. Adv. Neural Inf. Process. Syst.}, 2024.

\bibitem{qi2024shapellm}
Z.~Qi, R.~Dong, S.~Zhang, H.~Geng, C.~Han, Z.~Ge, L.~Yi, and K.~Ma, ``Shapellm: Universal 3d object understanding for embodied interaction,'' in \emph{Proc. Eur. Conf. Comput. Vis.}, 2024, pp. 214--238.

\bibitem{zha2023instance}
Y.~Zha, J.~Wang, T.~Dai, B.~Chen, Z.~Wang, and S.-T. Xia, ``Instance-aware dynamic prompt tuning for pre-trained point cloud models,'' in \emph{Proc. IEEE Int. Conf. Comput. Vis.}, 2023, pp. 14\,161--14\,170.

\bibitem{zhou2024dynamic}
X.~Zhou, D.~Liang, W.~Xu, X.~Zhu, Y.~Xu, Z.~Zou, and X.~Bai, ``Dynamic adapter meets prompt tuning: Parameter-efficient transfer learning for point cloud analysis,'' in \emph{Proc. IEEE Conf. Comput. Vis. Pattern Recognit.}, 2024, pp. 14\,707--14\,717.

\bibitem{tang2024point}
Y.~Tang, R.~Zhang, Z.~Guo, X.~Ma, B.~Zhao, Z.~Wang, D.~Wang, and X.~Li, ``Point-peft: Parameter-efficient fine-tuning for 3d pre-trained models,'' in \emph{Proc. AAAI Conf. Artif. Intell.}, vol.~38, no.~6, 2024, pp. 5171--5179.

\bibitem{fei2024fine}
J.~Fei and Z.~Deng, ``Fine-tuning point cloud transformers with dynamic aggregation,'' in \emph{Proc. IEEE Int. Conf. Robotics Automation}, 2024, pp. 9455--9462.

\bibitem{gao2020graphter}
X.~Gao, W.~Hu, and G.-J. Qi, ``Graphter: Unsupervised learning of graph transformation equivariant representations via auto-encoding node-wise transformations,'' in \emph{Proc. IEEE Conf. Comput. Vis. Pattern Recognit.}, 2020, pp. 7163--7172.

\bibitem{rue2005gaussian}
H.~Rue and L.~Held, \emph{Gaussian Markov random fields: theory and applications}.\hskip 1em plus 0.5em minus 0.4em\relax Chapman and Hall/CRC, 2005.

\bibitem{zhang2014point}
C.~Zhang, D.~Florencio, and C.~Loop, ``Point cloud attribute compression with graph transform,'' in \emph{Proc. Int. Conf. Image Process.}, 2014, pp. 2066--2070.

\bibitem{hu2021graph}
W.~Hu, J.~Pang, X.~Liu, D.~Tian, C.-W. Lin, and A.~Vetro, ``Graph signal processing for geometric data and beyond: Theory and applications,'' \emph{IEEE Trans. Multimedia}, vol.~24, pp. 3961--3977, 2021.

\bibitem{qi2017pointnet}
C.~R. Qi, H.~Su, K.~Mo, and L.~J. Guibas, ``Pointnet: Deep learning on point sets for 3d classification and segmentation,'' in \emph{Proc. IEEE Conf. Comput. Vis. Pattern Recognit.}, 2017, pp. 652--660.

\bibitem{qi2017pointnet++}
C.~R. Qi, L.~Yi, H.~Su, and L.~J. Guibas, ``Pointnet++: Deep hierarchical feature learning on point sets in a metric space,'' in \emph{Proc. Adv. Neural Inf. Process. Syst.}, vol.~30, 2017.

\bibitem{rao2020global}
Y.~Rao, J.~Lu, and J.~Zhou, ``Global-local bidirectional reasoning for unsupervised representation learning of 3d point clouds,'' in \emph{Proc. IEEE Conf. Comput. Vis. Pattern Recognit.}, 2020, pp. 5376--5385.

\bibitem{qian2022pointnext}
G.~Qian, Y.~Li, H.~Peng, J.~Mai, H.~Hammoud, M.~Elhoseiny, and B.~Ghanem, ``Pointnext: Revisiting pointnet++ with improved training and scaling strategies,'' in \emph{Proc. Adv. Neural Inf. Process. Syst.}, vol.~35, 2022, pp. 23\,192--23\,204.

\bibitem{lin2023meta}
H.~Lin, X.~Zheng, L.~Li, F.~Chao, S.~Wang, Y.~Wang, Y.~Tian, and R.~Ji, ``Meta architecture for point cloud analysis,'' in \emph{Proc. IEEE Conf. Comput. Vis. Pattern Recognit.}, 2023, pp. 17\,682--17\,691.

\bibitem{sun2024x}
S.~Sun, Y.~Rao, J.~Lu, and H.~Yan, ``X-3d: Explicit 3d structure modeling for point cloud recognition,'' in \emph{Proc. IEEE Conf. Comput. Vis. Pattern Recognit.}, 2024, pp. 5074--5083.

\bibitem{wang2024gpsformer}
C.~Wang, M.~Wu, S.-K. Lam, X.~Ning, S.~Yu, R.~Wang, W.~Li, and T.~Srikanthan, ``Gpsformer: A global perception and local structure fitting-based transformer for point cloud understanding,'' in \emph{Proc. Eur. Conf. Comput. Vis.}, 2024, pp. 75--92.

\bibitem{zhao2021point}
H.~Zhao, L.~Jiang, J.~Jia, P.~H. Torr, and V.~Koltun, ``Point transformer,'' in \emph{Proc. IEEE Int. Conf. Comput. Vis.}, 2021, pp. 16\,259--16\,268.

\bibitem{guo2021pct}
M.-H. Guo, J.-X. Cai, Z.-N. Liu, T.-J. Mu, R.~R. Martin, and S.-M. Hu, ``Pct: Point cloud transformer,'' \emph{Computational Visual Media}, vol.~7, pp. 187--199, 2021.

\bibitem{wu2022point}
X.~Wu, Y.~Lao, L.~Jiang, X.~Liu, and H.~Zhao, ``Point transformer v2: Grouped vector attention and partition-based pooling,'' in \emph{Proc. Adv. Neural Inf. Process. Syst.}, vol.~35, 2022, pp. 33\,330--33\,342.

\bibitem{wu2024point}
X.~Wu, L.~Jiang, P.-S. Wang, Z.~Liu, X.~Liu, Y.~Qiao, W.~Ouyang, T.~He, and H.~Zhao, ``Point transformer v3: Simpler faster stronger,'' in \emph{Proc. IEEE Conf. Comput. Vis. Pattern Recognit.}, 2024, pp. 4840--4851.

\bibitem{zhang2023flattening}
Q.~Zhang, J.~Hou, Y.~Qian, Y.~Zeng, J.~Zhang, and Y.~He, ``Flattening-net: Deep regular 2d representation for 3d point cloud analysis,'' \emph{IEEE Trans. Pattern Anal. Mach. Intell.}, vol.~45, no.~8, pp. 9726--9742, 2023.

\bibitem{zhang2023learning}
R.~Zhang, L.~Wang, Y.~Qiao, P.~Gao, and H.~Li, ``Learning 3d representations from 2d pre-trained models via image-to-point masked autoencoders,'' in \emph{Proc. IEEE Conf. Comput. Vis. Pattern Recognit.}, 2023, pp. 21\,769--21\,780.

\bibitem{qi2023contrast}
Z.~Qi, R.~Dong, G.~Fan, Z.~Ge, X.~Zhang, K.~Ma, and L.~Yi, ``Contrast with reconstruct: Contrastive 3d representation learning guided by generative pretraining,'' in \emph{Proc. Int. Conf. Mach. Learn.}, 2023, pp. 28\,223--28\,243.

\bibitem{xie2020pointcontrast}
S.~Xie, J.~Gu, D.~Guo, C.~R. Qi, L.~Guibas, and O.~Litany, ``Pointcontrast: Unsupervised pre-training for 3d point cloud understanding,'' in \emph{Proc. Eur. Conf. Comput. Vis.}, 2020, pp. 574--591.

\bibitem{zheng2024point}
X.~Zheng, X.~Huang, G.~Mei, Y.~Hou, Z.~Lyu, B.~Dai, W.~Ouyang, and Y.~Gong, ``Point cloud pre-training with diffusion models,'' in \emph{Proc. IEEE Conf. Comput. Vis. Pattern Recognit.}, 2024, pp. 22\,935--22\,945.

\bibitem{feng2024shape2scene}
T.~Feng, W.~Wang, R.~Quan, and Y.~Yang, ``Shape2scene: 3d scene representation learning through pre-training on shape data,'' in \emph{Proc. Eur. Conf. Comput. Vis.}, 2024, pp. 73--91.

\bibitem{afham2022crosspoint}
M.~Afham, I.~Dissanayake, D.~Dissanayake, A.~Dharmasiri, K.~Thilakarathna, and R.~Rodrigo, ``Crosspoint: Self-supervised cross-modal contrastive learning for 3d point cloud understanding,'' in \emph{Proc. IEEE Conf. Comput. Vis. Pattern Recognit.}, 2022, pp. 9902--9912.

\bibitem{chen2023clip2scene}
R.~Chen, Y.~Liu, L.~Kong, X.~Zhu, Y.~Ma, Y.~Li, Y.~Hou, Y.~Qiao, and W.~Wang, ``Clip2scene: Towards label-efficient 3d scene understanding by clip,'' in \emph{Proc. IEEE Conf. Comput. Vis. Pattern Recognit.}, 2023, pp. 7020--7030.

\bibitem{chen2022adaptformer}
S.~Chen, C.~Ge, Z.~Tong, J.~Wang, Y.~Song, J.~Wang, and P.~Luo, ``Adaptformer: Adapting vision transformers for scalable visual recognition,'' in \emph{Proc. Adv. Neural Inf. Process. Syst.}, vol.~35, 2022, pp. 16\,664--16\,678.

\bibitem{sung2022lst}
Y.-L. Sung, J.~Cho, and M.~Bansal, ``Lst: Ladder side-tuning for parameter and memory efficient transfer learning,'' in \emph{Proc. Adv. Neural Inf. Process. Syst.}, vol.~35, 2022, pp. 12\,991--13\,005.

\bibitem{zhang2023adaptive}
Q.~Zhang, M.~Chen, A.~Bukharin, P.~He, Y.~Cheng, W.~Chen, and T.~Zhao, ``Adaptive budget allocation for parameter-efficient fine-tuning,'' in \emph{Proc. Int. Conf. Learn. Representations}, 2023.

\bibitem{shi2024dept}
Z.~Shi and A.~Lipani, ``Dept: Decomposed prompt tuning for parameter-efficient fine-tuning,'' in \emph{Proc. Int. Conf. Learn. Representations}, 2024.

\bibitem{ding2023parameter}
N.~Ding, Y.~Qin, G.~Yang, F.~Wei, Z.~Yang, Y.~Su, S.~Hu, Y.~Chen, C.-M. Chan, W.~Chen \emph{et~al.}, ``Parameter-efficient fine-tuning of large-scale pre-trained language models,'' \emph{Nature Mach. Intell.}, vol.~5, no.~3, pp. 220--235, 2023.

\bibitem{jia2022visual}
M.~Jia, L.~Tang, B.-C. Chen, C.~Cardie, S.~Belongie, B.~Hariharan, and S.-N. Lim, ``Visual prompt tuning,'' in \emph{Proc. Eur. Conf. Comput. Vis.}, 2022, pp. 709--727.

\bibitem{lian2022scaling}
D.~Lian, D.~Zhou, J.~Feng, and X.~Wang, ``Scaling \& shifting your features: A new baseline for efficient model tuning,'' in \emph{Proc. Adv. Neural Inf. Process. Syst.}, vol.~35, 2022, pp. 109--123.

\bibitem{zaken2022bitfit}
E.~B. Zaken, Y.~Goldberg, and S.~Ravfogel, ``Bitfit: Simple parameter-efficient fine-tuning for transformer-based masked language-models,'' in \emph{Proc. Annual Meeting of the Association for Computational Linguistics}, 2022, pp. 1--9.

\bibitem{tu2023visual}
C.-H. Tu, Z.~Mai, and W.-L. Chao, ``Visual query tuning: Towards effective usage of intermediate representations for parameter and memory efficient transfer learning,'' in \emph{Proc. IEEE Conf. Comput. Vis. Pattern Recognit.}, 2023, pp. 7725--7735.

\bibitem{houlsby2019parameter}
N.~Houlsby, A.~Giurgiu, S.~Jastrzebski, B.~Morrone, Q.~De~Laroussilhe, A.~Gesmundo, M.~Attariyan, and S.~Gelly, ``Parameter-efficient transfer learning for nlp,'' in \emph{Proc. Int. Conf. Mach. Learn.}, 2019, pp. 2790--2799.

\bibitem{hu2021lora}
E.~J. Hu, P.~Wallis, Z.~Allen-Zhu, Y.~Li, S.~Wang, L.~Wang, W.~Chen \emph{et~al.}, ``Lora: Low-rank adaptation of large language models,'' in \emph{Proc. Int. Conf. Learn. Representations}, 2022.

\bibitem{li2024adapter}
M.~Li, P.~Ye, Y.~Huang, L.~Zhang, T.~Chen, T.~He, J.~Fan, and W.~Ouyang, ``Adapter-x: A novel general parameter-efficient fine-tuning framework for vision,'' \emph{arXiv:2406.03051}, 2024.

\bibitem{lester2021power}
B.~Lester, R.~Al-Rfou, and N.~Constant, ``The power of scale for parameter-efficient prompt tuning,'' in \emph{Proc. Conf. Empirical Methods in Natural Language Process.}, 2021, pp. 3045--3059.

\bibitem{li2021prefix}
X.~L. Li and P.~Liang, ``Prefix-tuning: Optimizing continuous prompts for generation,'' in \emph{Proc. Annual Meeting of the Association for Computational Linguistics}, 2021, pp. 4582--4597.

\bibitem{li2024adapt}
M.~Li, D.~Li, G.~Yang, Y.-m. Cheung, and H.~Huang, ``Adapt pointformer: 3d point cloud analysis via adapting 2d visual transformers,'' in \emph{Proc. Eur. Conf. Artif. Intell.}\hskip 1em plus 0.5em minus 0.4em\relax IOS Press, 2024, pp. 89--96.

\bibitem{tang2024any2point}
Y.~Tang, R.~Zhang, J.~Liu, Z.~Guo, B.~Zhao, Z.~Wang, P.~Gao, H.~Li, D.~Wang, and X.~Li, ``Any2point: Empowering any-modality large models for efficient 3d understanding,'' in \emph{Proc. Eur. Conf. Comput. Vis.}, 2024, pp. 456--473.

\bibitem{sun24ppt}
H.~Sun, Y.~Wang, W.~Chen, H.~Deng, and D.~Li, ``Parameter-efficient prompt learning for 3d point cloud understanding,'' in \emph{Proc. IEEE Int. Conf. Robotics Automation}, 2024, pp. 9478--9486.

\bibitem{fei2024parameter}
B.~Fei, L.~Liu, W.~Yang, Z.~Li, W.-M. Chen, and L.~Ma, ``Parameter efficient point cloud prompt tuning for unified point cloud understanding,'' \emph{IEEE Trans. Intell. Vehicles}, 2024.

\bibitem{wang2019dynamic}
Y.~Wang, Y.~Sun, Z.~Liu, S.~E. Sarma, M.~M. Bronstein, and J.~M. Solomon, ``Dynamic graph cnn for learning on point clouds,'' \emph{ACM Trans. ON Graphics}, vol.~38, no.~5, pp. 1--12, 2019.

\bibitem{te2018rgcnn}
G.~Te, W.~Hu, A.~Zheng, and Z.~Guo, ``Rgcnn: Regularized graph cnn for point cloud segmentation,'' in \emph{Proc. ACM Multimedia}, 2018, pp. 746--754.

\bibitem{chen2018pointagcn}
L.~Chen, G.~Wei, and Z.~Wang, ``Pointagcn: Adaptive spectral graph cnn for point cloud feature learning,'' in \emph{Proc. of Int. Conf. on Security, Pattern Anal., and Cybernetics}, 2018, pp. 401--406.

\bibitem{li2019deepgcns}
G.~Li, M.~Muller, A.~Thabet, and B.~Ghanem, ``Deepgcns: Can gcns go as deep as cnns?'' in \emph{Proc. IEEE Int. Conf. Comput. Vis.}, 2019, pp. 9267--9276.

\bibitem{lu2020pointngcnn}
Q.~Lu, C.~Chen, W.~Xie, and Y.~Luo, ``Pointngcnn: Deep convolutional networks on 3d point clouds with neighborhood graph filters,'' \emph{Computers \& Graphics}, vol.~86, pp. 42--51, 2020.

\bibitem{wen2024pointwavelet}
C.~Wen, J.~Long, B.~Yu, and D.~Tao, ``Pointwavelet: Learning in spectral domain for 3-d point cloud analysis,'' \emph{IEEE Trans. Neural Networks Learn. Syst.}, vol.~36, no.~3, pp. 4400--4412, 2024.

\bibitem{wang2018local}
C.~Wang, B.~Samari, and K.~Siddiqi, ``Local spectral graph convolution for point set feature learning,'' in \emph{Proc. Eur. Conf. Comput. Vis.}, 2018, pp. 52--66.

\bibitem{zhang2020hypergraph}
S.~Zhang, S.~Cui, and Z.~Ding, ``Hypergraph spectral analysis and processing in 3d point cloud,'' \emph{IEEE Trans. Image Process.}, vol.~30, pp. 1193--1206, 2020.

\bibitem{liu2023point}
D.~Liu, W.~Hu, and X.~Li, ``Point cloud attacks in graph spectral domain: When 3d geometry meets graph signal processing,'' \emph{IEEE Trans. Pattern Anal. Mach. Intell.}, vol.~46, no.~5, pp. 3079--3095, 2023.

\bibitem{ramasinghe2020spectral}
S.~Ramasinghe, S.~Khan, N.~Barnes, and S.~Gould, ``Spectral-gans for high-resolution 3d point-cloud generation,'' in \emph{Proc. IEEE Int. Conf. Intell. Robots Syst.}, 2020, pp. 8169--8176.

\bibitem{yu2024visual}
B.~X. Yu, J.~Chang, H.~Wang, L.~Liu, S.~Wang, Z.~Wang, J.~Lin, L.~Xie, H.~Li, Z.~Lin \emph{et~al.}, ``Visual tuning,'' \emph{ACM Computing Surveys}, vol.~56, no.~12, pp. 1--38, 2024.

\bibitem{shuman2013emerging}
D.~I. Shuman, S.~K. Narang, P.~Frossard, A.~Ortega, and P.~Vandergheynst, ``The emerging field of signal processing on graphs: Extending high-dimensional data analysis to networks and other irregular domains,'' \emph{IEEE Signal Process. Mag.}, vol.~30, no.~3, pp. 83--98, 2013.

\bibitem{chung1997spectral}
F.~R. Chung, \emph{Spectral graph theory}.\hskip 1em plus 0.5em minus 0.4em\relax American Mathematical Soc., 1997, vol.~92.

\bibitem{ramachandran2017searching}
P.~Ramachandran, B.~Zoph, and Q.~V. Le, ``Searching for activation functions,'' in \emph{Proc. Int. Conf. Learn. Representations Workshops}, 2018.

\bibitem{uy2019revisiting}
M.~A. Uy, Q.-H. Pham, B.-S. Hua, T.~Nguyen, and S.-K. Yeung, ``Revisiting point cloud classification: A new benchmark dataset and classification model on real-world data,'' in \emph{Proc. IEEE Int. Conf. Comput. Vis.}, 2019, pp. 1588--1597.

\bibitem{wu20153d}
Z.~Wu, S.~Song, A.~Khosla, F.~Yu, L.~Zhang, X.~Tang, and J.~Xiao, ``3d shapenets: A deep representation for volumetric shapes,'' in \emph{Proc. IEEE Conf. Comput. Vis. Pattern Recognit.}, 2015, pp. 1912--1920.

\bibitem{morton1966computer}
G.~M. Morton, ``A computer oriented geodetic data base and a new technique in file sequencing,'' \emph{Physics of Plasmas}, 1966.

\bibitem{loshchilov2019decoupled}
I.~Loshchilov and F.~Hutter, ``Decoupled weight decay regularization,'' in \emph{Proc. Int. Conf. Learn. Representations}, 2017.

\bibitem{loshchilov2017sgdr}
------, ``Sgdr: Stochastic gradient descent with warm restarts,'' in \emph{Proc. Int. Conf. Learn. Representations}, 2017.

\bibitem{yi2016scalable}
L.~Yi, V.~G. Kim, D.~Ceylan, I.-C. Shen, M.~Yan, H.~Su, C.~Lu, Q.~Huang, A.~Sheffer, and L.~Guibas, ``A scalable active framework for region annotation in 3d shape collections,'' \emph{ACM Trans. ON Graphics}, vol.~35, no.~6, pp. 1--12, 2016.

\bibitem{armeni20163d}
I.~Armeni, O.~Sener, A.~R. Zamir, H.~Jiang, I.~Brilakis, M.~Fischer, and S.~Savarese, ``3d semantic parsing of large-scale indoor spaces,'' in \emph{Proc. IEEE Conf. Comput. Vis. Pattern Recognit.}, 2016, pp. 1534--1543.

\bibitem{tchapmi2017segcloud}
L.~Tchapmi, C.~Choy, I.~Armeni, J.~Gwak, and S.~Savarese, ``Segcloud: Semantic segmentation of 3d point clouds,'' in \emph{Proc. of Int. Conf. 3D Vis.}, 2017, pp. 537--547.

\bibitem{dai2017scannet}
A.~Dai, A.~X. Chang, M.~Savva, M.~Halber, T.~Funkhouser, and M.~Nie{\ss}ner, ``Scannet: Richly-annotated 3d reconstructions of indoor scenes,'' in \emph{Proc. IEEE Conf. Comput. Vis. Pattern Recognit.}, 2017, pp. 5828--5839.

\bibitem{yuan2018pcn}
W.~Yuan, T.~Khot, D.~Held, C.~Mertz, and M.~Hebert, ``Pcn: Point completion network,'' in \emph{Proc. of Int. Conf. 3D Vis.}, 2018, pp. 728--737.

\bibitem{hamdi2021mvtn}
A.~Hamdi, S.~Giancola, and B.~Ghanem, ``Mvtn: Multi-view transformation network for 3d shape recognition,'' in \emph{Proc. IEEE Int. Conf. Comput. Vis.}, 2021, pp. 1--11.

\bibitem{lin2021learning}
Z.-H. Lin, S.-Y. Huang, and Y.-C.~F. Wang, ``Learning of 3d graph convolution networks for point cloud analysis,'' \emph{IEEE Trans. Pattern Anal. Mach. Intell.}, vol.~44, no.~8, pp. 4212--4224, 2021.

\bibitem{ran2022surface}
H.~Ran, J.~Liu, and C.~Wang, ``Surface representation for point clouds,'' in \emph{Proc. IEEE Conf. Comput. Vis. Pattern Recognit.}, 2022, pp. 18\,942--18\,952.

\bibitem{ma2022rethinking}
X.~Ma, C.~Qin, H.~You, H.~Ran, and Y.~Fu, ``Rethinking network design and local geometry in point cloud: A simple residual mlp framework,'' \emph{Proc. Int. Conf. Learn. Representations}, 2022.

\bibitem{hong2023attention}
C.-Y. Hong, Y.-Y. Chou, and T.-L. Liu, ``Attention discriminant sampling for point clouds,'' in \emph{Proc. IEEE Int. Conf. Comput. Vis.}, 2023, pp. 14\,429--14\,440.

\bibitem{wang2021unsupervised}
H.~Wang, Q.~Liu, X.~Yue, J.~Lasenby, and M.~J. Kusner, ``Unsupervised point cloud pre-training via occlusion completion,'' in \emph{Proc. IEEE Int. Conf. Comput. Vis.}, 2021, pp. 9782--9792.

\bibitem{liu2022masked}
H.~Liu, M.~Cai, and Y.~J. Lee, ``Masked discrimination for self-supervised learning on point clouds,'' in \emph{Proc. Eur. Conf. Comput. Vis.}, 2022, pp. 657--675.

\bibitem{zha2024towards}
Y.~Zha, H.~Ji, J.~Li, R.~Li, T.~Dai, B.~Chen, Z.~Wang, and S.-T. Xia, ``Towards compact 3d representations via point feature enhancement masked autoencoders,'' in \emph{Proc. AAAI Conf. Artif. Intell.}, vol.~38, no.~7, 2024, pp. 6962--6970.

\bibitem{chang2015shapenet}
A.~X. Chang, T.~Funkhouser, L.~Guibas, P.~Hanrahan, Q.~Huang, Z.~Li, S.~Savarese, M.~Savva, S.~Song, H.~Su \emph{et~al.}, ``Shapenet: An information-rich 3d model repository,'' \emph{arXiv preprint arXiv:1512.03012}, 2015.

\bibitem{misra2021end}
I.~Misra, R.~Girdhar, and A.~Joulin, ``An end-to-end transformer model for 3d object detection,'' in \emph{Proc. IEEE Int. Conf. Comput. Vis.}, 2021, pp. 2906--2917.

\bibitem{yu2023adapointr}
X.~Yu, Y.~Rao, Z.~Wang, J.~Lu, and J.~Zhou, ``Adapointr: Diverse point cloud completion with adaptive geometry-aware transformers,'' \emph{IEEE Trans. Pattern Anal. Mach. Intell.}, vol.~45, no.~12, pp. 14\,114--14\,130, 2023.

\bibitem{gao2024parameter}
Z.~Gao, Q.~Wang, A.~Chen, Z.~Liu, B.~Wu, L.~Chen, and J.~Li, ``Parameter-efficient fine-tuning with discrete fourier transform,'' in \emph{Proc. Int. Conf. Mach. Learn.}, 2024.

\bibitem{jie2023fact}
S.~Jie and Z.-H. Deng, ``Fact: Factor-tuning for lightweight adaptation on vision transformer,'' in \emph{Proc. AAAI Conf. Artif. Intell.}, vol.~37, no.~1, 2023, pp. 1060--1068.

\bibitem{jie2023revisiting}
S.~Jie, H.~Wang, and Z.-H. Deng, ``Revisiting the parameter efficiency of adapters from the perspective of precision redundancy,'' in \emph{Proc. IEEE Int. Conf. Comput. Vis.}, 2023, pp. 17\,217--17\,226.

\bibitem{zhao2023sct}
H.~H. Zhao, P.~Wang, Y.~Zhao, H.~Luo, F.~Wang, and M.~Z. Shou, ``Sct: A simple baseline for parameter-efficient fine-tuning via salient channels,'' \emph{Int. J. Comput. Vis.}, vol. 132, no.~3, pp. 731--749, 2024.

\bibitem{van2008visualizing}
L.~Van~der Maaten and G.~Hinton, ``Visualizing data using t-sne,'' \emph{J. of Mach. learn. research}, vol.~9, no.~11, 2008.

\bibitem{von2007tutorial}
U.~Von~Luxburg, ``A tutorial on spectral clustering,'' \emph{Statistics and computing}, vol.~17, pp. 395--416, 2007.

\end{thebibliography}
}

\end{document}